\pdfoutput=1

\documentclass[journal]{IEEEtran}
\usepackage{cite}
\ifCLASSINFOpdf
   \usepackage[pdftex]{graphicx}
\else
   \usepackage[dvips]{graphicx}
\fi
\usepackage{amsmath}
\usepackage{array}
\ifCLASSOPTIONcompsoc
  \usepackage[caption=false,font=normalsize,labelfont=sf,textfont=sf]{subfig}
\else
  \usepackage[caption=false,font=footnotesize]{subfig}
\fi
\usepackage{dblfloatfix}
\ifCLASSOPTIONcaptionsoff
  \usepackage[nomarkers]{endfloat}
 \let\MYoriglatexcaption\caption
 \renewcommand{\caption}[2][\relax]{\MYoriglatexcaption[#2]{#2}}
\fi
\usepackage{url}
\usepackage{color}% The package I import.
\usepackage{booktabs}% The package I import.

\usepackage{algorithm}% The package I import.
\usepackage{algpseudocode}% The package I import.
\makeatletter % The package I import.
\newcommand{\Rmnum}[1]{\expandafter\@slowromancap\romannumeral #1@} % The package I import.
\makeatother % The package I import.

\usepackage{tikz} % The package I import.
\usepackage{pgfplots} % The package I import.
\usepackage[T1]{fontenc} % The package I import
\usepackage{xcolor} % The package I import {red/blue/green/black/white/cyan/magenta/yellow}

\definecolor{PROTOSS PYLON}{RGB}{0, 168, 255} % The package I import.
\definecolor{VANADYL BLUE}{RGB}{0, 151, 230} % The package I import.

\definecolor{PERIWINKLE}{RGB}{156, 136, 255} % The package I import.
\definecolor{MATT PURPLE}{RGB}{140, 122, 230} % The package I import.

\definecolor{RISE-N-SHINE}{RGB}{251, 197, 49} % The package I import.
\definecolor{NANOHANACHA GOLD}{RGB}{225, 177, 44} % The package I import.

\definecolor{DOWNLOAD PROGRESS}{RGB}{76, 209, 55} % The package I import.
\definecolor{SKIRRET GREEN}{RGB}{68, 189, 50} % The package I import.

\definecolor{SEABROOK}{RGB}{72, 126, 176} % The package I import.
\definecolor{NAVAL}{RGB}{64, 115, 158} % The package I import.

\definecolor{NASTURCIAN FLOWER}{RGB}{232, 65, 24} % The package I import.
\definecolor{HARLEY DAVIDSON ORANGE}{RGB}{194, 54, 22} % The package I import.

\definecolor{LYNX WHITE}{RGB}{245, 246, 250} % The package I import.
\definecolor{HINT OF PENSIVE}{RGB}{220, 221, 225} % The package I import.

\definecolor{BLUEBERRY SODA}{RGB}{127, 143, 166} % The package I import.
\definecolor{CHAIN GANG GREY}{RGB}{113, 128, 147} % The package I import.

\definecolor{MAZARINE BLUE}{RGB}{39, 60, 117} % The package I import.
\definecolor{PICO VOID}{RGB}{25, 42, 86} % The package I import.

\definecolor{BLUE NIGHTS}{RGB}{53, 59, 72} % The package I import.
\definecolor{ELECTROMAGNETIC}{RGB}{47,54,64} % The package I import.

\begin{document}
\onecolumn
\copyright 20XX IEEE.  Personal use of this material is permitted.  Permission from IEEE must be obtained for all other uses, in any current or future media, including reprinting/republishing this material for advertising or promotional purposes, creating new collective works, for resale or redistribution to servers or lists, or reuse of any copyrighted component of this work in other works.
\twocolumn

%
% paper title
% Titles are generally capitalized except for words such as a, an, and, as,
% at, but, by, for, in, nor, of, on, or, the, to and up, which are usually
% not capitalized unless they are the first or last word of the title.
% Linebreaks \\ can be used within to get better formatting as desired.
% Do not put math or special symbols in the title.

\title{Decoder Choice Network for Meta-Learning}
%\title{Multi Heads Parameter Mapping for Meta-Learning}

%
%
% author names and IEEE memberships
% note positions of commas and nonbreaking spaces ( ~ ) LaTeX will not break
% a structure at a ~ so this keeps an author's name from being broken across
% two lines.
% use \thanks{} to gain access to the first footnote area
% a separate \thanks must be used for each paragraph as LaTeX2e's \thanks
% was not built to handle multiple paragraphs
%

\author{Jialin~Liu,
Fei~Chao,~\IEEEmembership{Member,~IEEE,}
        Longzhi~Yang,~\IEEEmembership{Senior Member,~IEEE,}
        Chih-Min~Lin,~\IEEEmembership{Fellow,~IEEE,}
        and~Qiang~Shen % <-this % stops a space
\thanks{J. Liu and F. Chao are with the Department of Artificial Intelligence, School of Informatics, Xiamen University, China e-mail: (fchao@xmu.edu.cn). L. Yang is with the Department of Computer and Information Sciences, Northumbria University, UK. C.-M. Lin is with the Department of Electrical Engineering, Yuan Ze University, Taiwan. F. Chao and Q. Shen are with Institute of Mathematics, Physics and Computer Science, Aberystwyth University, UK. Corresponding Author: Fei Chao}% <-this % stops a space
% <-this % stops a space
\thanks{This work was supported by the National Natural Science Foundation of China (No. 61673322, 61673326, and 91746103), Fundamental Research Funds for the Central Universities (No. 20720190142), Natural Science Foundation of Fujian Province of China (No. 2017J01128 and 2017J01129), and European Union's Horizon 2020 research and innovation programme under the Marie Sklodowska-Curie grant agreement No. 663830.}% <-this % stops a space
\thanks{Manuscript received; revised.}}

% note the % following the last \IEEEmembership and also \thanks -
% these prevent an unwanted space from occurring between the last author name
% and the end of the author line. i.e., if you had this:
%
% \author{....lastname \thanks{...} \thanks{...} }
%                     ^------------^------------^----Do not want these spaces!
%
% a space would be appended to the last name and could cause every name on that
% line to be shifted left slightly. This is one of those "LaTeX things". For
% instance, "\textbf{A} \textbf{B}" will typeset as "A B" not "AB". To get
% "AB" then you have to do: "\textbf{A}\textbf{B}"
% \thanks is no different in this regard, so shield the last } of each \thanks
% that ends a line with a % and do not let a space in before the next \thanks.
% Spaces after \IEEEmembership other than the last one are OK (and needed) as
% you are supposed to have spaces between the names. For what it is worth,
% this is a minor point as most people would not even notice if the said evil
% space somehow managed to creep in.

% The paper headers
\markboth{Journal of \LaTeX\ Class Files,~Vol.~14, No.~8, August~2015}%
{Shell \MakeLowercase{\textit{et al.}}: Bare Demo of IEEEtran.cls for IEEE Journals}
% The only time the second header will appear is for the odd numbered pages
% after the title page when using the twoside option.
%
% *** Note that you probably will NOT want to include the author's ***
% *** name in the headers of peer review papers.                   ***
% You can use \ifCLASSOPTIONpeerreview for conditional compilation here if
% you desire.

% If you want to put a publisher's ID mark on the page you can do it like
% this:
%\IEEEpubid{0000--0000/00\$00.00~\copyright~2015 IEEE}
% Remember, if you use this you must call \IEEEpubidadjcol in the second
% column for its text to clear the IEEEpubid mark.

% use for special paper notices
%\IEEEspecialpapernotice{(Invited Paper)}

% make the title area
\maketitle

% As a general rule, do not put math, special symbols or citations
% in the abstract or keywords.
\begin{abstract}
   Meta-learning has been widely used for implementing few-shot learning and fast model adaptation. One kind of meta-learning methods attempt to learn how to control the gradient descent process in order to make the gradient-based learning have high speed and generalization. This work proposes a method that controls the gradient descent process of the model parameters of a neural network by limiting the model parameters in a low-dimensional latent space. The main challenge of this idea is that a decoder with too many parameters is required. This work designs a decoder with typical structure and shares a part of weights in the decoder to reduce the number of the required parameters. Besides, this work has introduced ensemble learning to work with the proposed approach for improving performance. The results show that the proposed approach is witnessed by the superior performance over the Omniglot classification and the miniImageNet classification tasks.
   %Different latent spaces are chosen for different tasks in order to increase the generalization performance of gradient-based learning. This process is implemented by choosing the different decoder networks based on the task features.
\end{abstract}

% Note that keywords are not normally used for peerreview papers.

% For peer review papers, you can put extra information on the cover
% page as needed:
% \ifCLASSOPTIONpeerreview
% \begin{center} \bfseries EDICS Category: 3-BBND \end{center}
% \fi
%
% For peerreview papers, this IEEEtran command inserts a page break and
% creates the second title. It will be ignored for other modes.
\IEEEpeerreviewmaketitle

\section{Introduction}\label{Introduction}
% The very first letter is a 2 line initial drop letter followed
% by the rest of the first word in caps.
%
% form to use if the first word consists of a single letter:
% \IEEEPARstart{A}{demo} file is ....
%
% form to use if you need the single drop letter followed by
% normal text (unknown if ever used by the IEEE):
% \IEEEPARstart{A}{}demo file is ....
%
% Some journals put the first two words in caps:
% \IEEEPARstart{T}{his demo} file is ....
%
% Here we have the typical use of a "T" for an initial drop letter
% and "HIS" in caps to complete the first word.
\IEEEPARstart{M}{achine} learning has recently demonstrated near-human performance in the traditionally challenging tasks of object recognition, image classification, and games and scenario generation, amongst other applications. The key to such successes is the availability or obtainability of high-quality large datasets. Collecting and labeling data or harvesting labeled data from literature and historic archives require significant human efforts, but the resulted dataset can usually only be used for one specific task. % by {\color{orange}conventional machine learning approaches}.
However, humans have the ability to quickly learn new conceptions and skills for novel tasks based on prior knowledge and experience; meta-learning is a branch of machine learning techniques imitating such ability by learning parameters fine-tuning from prior datasets and pre-training models. Consequently, meta-learning can extend the boundary of machine learning greatly, which concerns the distributions of tasks in addition to the traditionally-used distribution of data samples. It not only enables the `reuse' of datasets across different tasks, but also prevents from over-fitting to new, and usually small dataset for novel tasks~\cite{Li}. In this case, each novel task is a learning task, which is supported by training examples (or shot) and testing examples (or query)~\cite{Shedivat}.

The most widely studied form of meta-learning is few-shot learning, which requires models to predict labels of instances from unseen classes during the testing phase, with the support of only a few labeled samples from each category. Many methods have recently been proposed to implement few-shot learning tasks, which can be categorized into three types \cite{Andrei}: memory-based, optimization-based, and metric-based. The memory-based methods extend a memory space to store key training examples or model-related information, and it is often achieved by using attention model~\cite{Santoro,Tsendsuren}. Optimization-based methods learn to control the process of optimization within each task, by learning the initial parameters (e.g.,\cite{Chelsea,Andrei,Alex}) or the optimizer (e.g.,\cite{Sachin,Fengwei}). Metric-based methods focus on learning similarity metrics which maximizing the similarity between members from the same class. %These options provide the flexibility for wide application domains.

The purpose of this work is to design an optimization-based method, which controls the optimization of model parameters by limiting those parameters in a low dimension space. This inspiration comes from neural style transfer~\cite{bansal2018can}, which updates the input image of a deep network rather than parameters. If we replace the input image as the latent code and the output as the model parameters, we can indirectly update the model parameters by updating the latent code. The network maps the latent code to the model parameters is called decoder in this work.

However, this idea is difficult to implement for the high-dimension model parameter space. If the fully connected network is adopted as the decoder, the complexity of the decoder is usually square of the number of the model parameters. Instead of a fully connected network, we propose a new structure, named group linear transformation (GLT), with lower time and space complexity to denote the decoder network (detail in \ref{Weight_Sharing}).

Besides, we also draw on the idea of task-dependent adaptive metric (TADAM)~\cite{Oreshkin} and latent embedding optimization (LEO)~\cite{Andrei}, which enhances the correlation between the model and the task by making the model parameters depending on the task. We enhances the correlation between the decoder and the task by choosing the different decoders based on the task features by a choice network. Due to all decoders are able to share a part of their parameters and the complexity of the choice network is lower than that of the decoders, this approach makes the decoders be task-dependent with few costs.

Finally, in order to further improve generalization of the model, this work adopts the training protocol proposed in snapshot ensemble~\cite{Huang} instead of the standard training protocol. The training protocol with snapshot ensemble~\cite{Huang} selects the models with the highest accuracy on the validation set in a single training process to enable the ensemble process~\cite{Huang}.

We evaluate DCN on both regression and classification few-shot learning tasks. The experiment results show the proposed optimization-based method greatly improves the accuracy of the few-shot learning model by enhancing the dependency between model and task, and even have learnable fewer parameters in several tasks. The main contribution of this work is threefold: 1) an efficient structure which enables the gradient control for the high-dimension model parameters by the low-dimension space, 2) the implementation of the tack-dependent gradient control by choosing the decoders based on the task features, and 3) the integration of snapshot ensemble in the proposed DCN for performance improvement.

The rest of the paper is organized as follows. Section~\ref{Sec_Background} reviews the underpinning theory of the proposed work. Section~\ref{Sec_FPMR} details the proposed DCN. Section~\ref{Sec_Experimentation} reports the results. Section~\ref{Sec_Conclusion} concludes the work with a list of future work provided.

\section{Background}\label{Sec_Background}
The theoretical underpinning of the proposed approach is reviewed in this section, including few-shot learning and meta-learning.

\subsection{Few-Shot Learning}
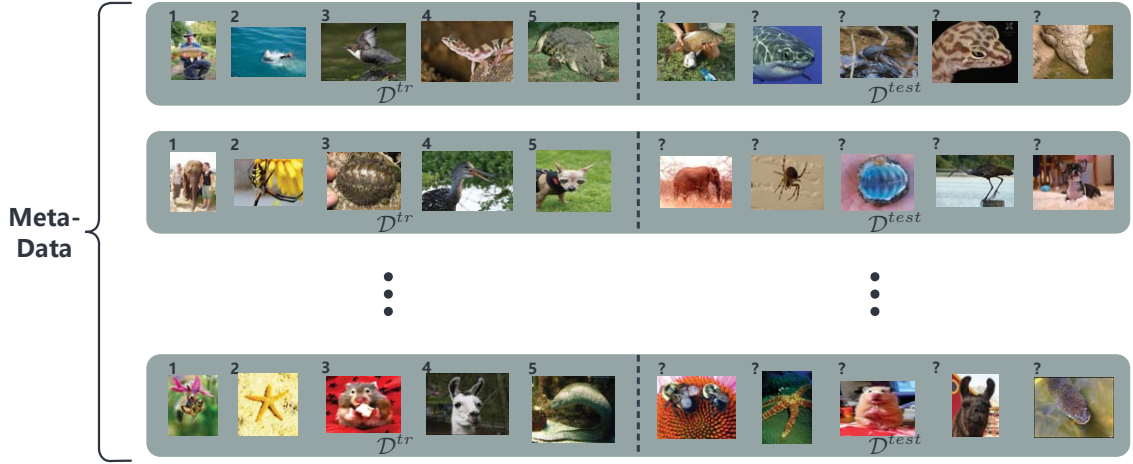
\begin{figure*}
\begin{center}
\includegraphics[width=150mm]{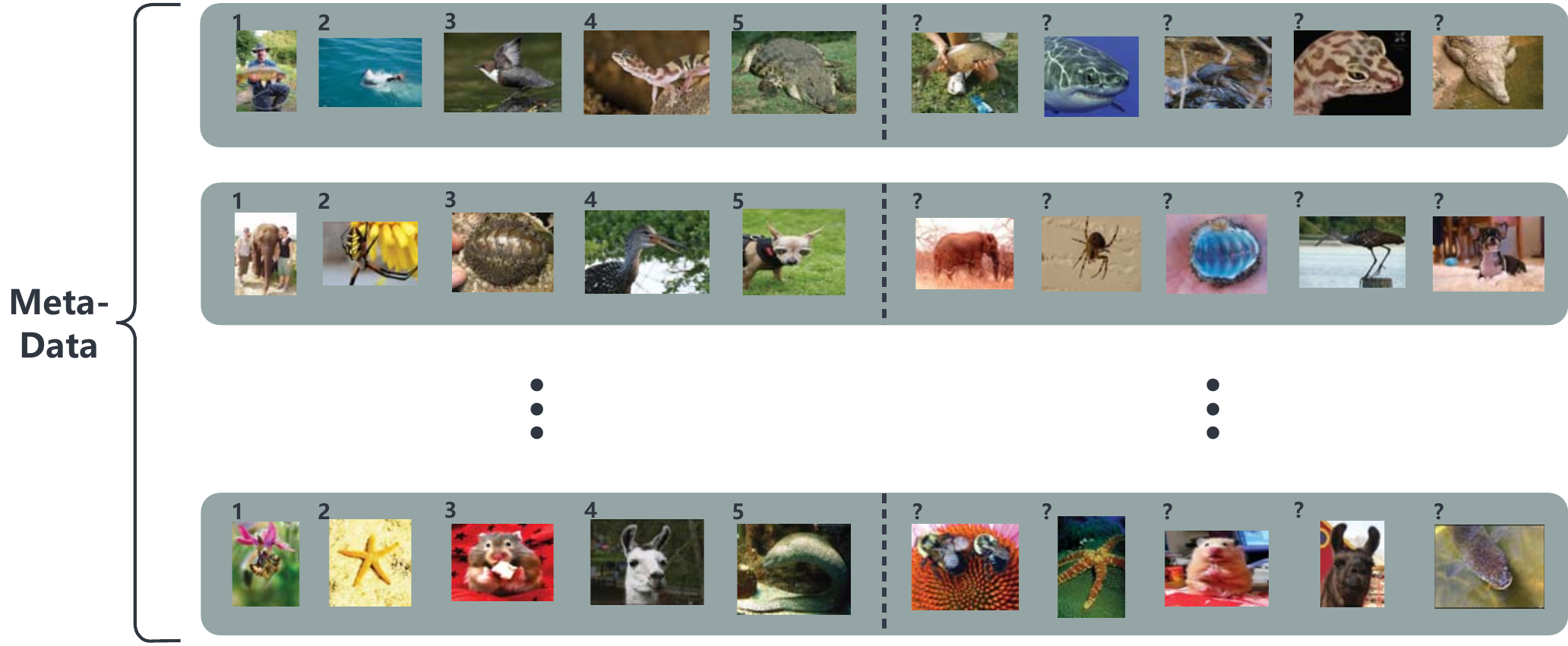}
    \put(-285,137){\small \color{ELECTROMAGNETIC} $\mathcal{D}^{tr}$}
    \put(-99,137){\small \color{ELECTROMAGNETIC} $\mathcal{D}^{test}$}
    \put(-285,88){\small \color{ELECTROMAGNETIC} $\mathcal{D}^{tr}$}
    \put(-99,88){\small \color{ELECTROMAGNETIC} $\mathcal{D}^{test}$}
    \put(-285,4){\small \color{ELECTROMAGNETIC} $\mathcal{D}^{tr}$}
    \put(-99,4){\small \color{ELECTROMAGNETIC} $\mathcal{D}^{test}$}
\end{center}
   \caption{Example of few-shot learning data. These are the instances from a 5-way, 1-shot, 1-query meta-data. Each few learning task contains ten images from different classes, each class has one training example and one testing example. In the figure, images with the same number in the upper left corner of each line are from the same class.}
\label{MetaData}
\end{figure*}

%Few-shot learning differs from {\color{orange}conventional machine learning} in that the object of learning is not a specific task trained by a dataset with input features and output features, but between tasks.
%{\color{orange}[cut]}

In supervised learning, the training dataset is  a number of labelled data instances $\mathcal{D}=\{({\rm x}^1,{\rm y}^1),({\rm x}^2,{\rm y}^2),\cdots, ({\rm x}^K,{\rm y}^K)\}$, with each (${\rm x}^i, {\rm y}^i$), $1\leq i\leq K$ being a data instance with the features ${\rm x}^i$ and the labels ${\rm y}^i$, where $K$ is the number of data instances. Differently, few-shot learning learns between tasks, and thus the input dataset can be represented as $\mathcal{D}^{meta}=\{\mathcal{D}^{tr}_i,\mathcal{D}^{test}_i\}_i$, where $\mathcal{D}^{tr}_i=\{{\rm x}^{tr}_{ij}, {\rm y}^{tr}_{ij}\}_j$ and $\mathcal{D}^{test}_i=\{{\rm x}^{test}_{ij}, {\rm y}^{test}_{ij}\}_j$. In other words, few-shot learning takes each dataset regarding tasks as instances of training.

We explain the definition of N-way, $\rm K$-shot H-query tasks. It means those tasks are N classification tasks, each with $\rm K$ training examples and H testing examples~\cite{Oriol}. Fig.~\ref{MetaData} shows the 5-way, 1-shot, 1-query task of miniImageNet, and each class in a task has 1 training instances and 1 testing instances. In the 5-way, 5-shot, 1-query task, there are 5 training instances and 5 testing instances in each class.
%An example of {\color{orange}5-way, 1-shot, 1-query task} of miniImageNet is illustrated in Fig.~\ref{MetaData}.

The objectives of supervised learning tasks and few-shot learning tasks can be expressed by the following optimization tasks respectively:
\begin{equation}
\label{regular}
\hat{\theta}=\mathop{\arg\min}_\theta\sum_i\mathcal{L}(f_\theta({\rm x}^i),{\rm y}^i),
\end{equation}
\begin{equation}
\label{few-shot}
\hat{\theta}=\mathop{\arg\min}_\theta\sum_i\sum_j\mathcal{L}(f_{\{\mathcal{D}^{tr}_i,\theta\}}({\rm x}^{test}_{ij}),{\rm y}^{test}_{ij})
\end{equation}
Eq. (\ref{regular}) represents a standard empirical risk minimization task of supervised learning, in which the prediction of ${\rm y}^i$ only depends on ${\rm x}^i$ and $\theta$, but when it comes to few-shot learning as shown in Eq. (\ref{few-shot}), the prediction of ${\rm y}_{ij}^{test}$ also depends on the training examples of $\mathcal{D}^{tr}_i$, in addition to the corresponding the features ${\rm x}^{test}_{ij}$ and the parameters $\theta$.

\subsection{Meta-Learning}

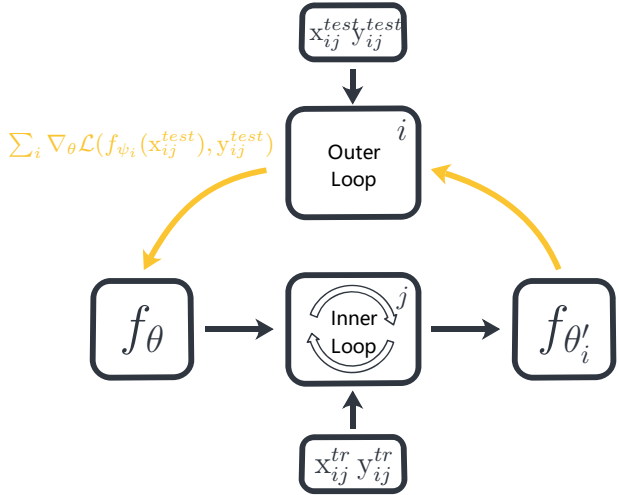
\begin{figure}[t]
\begin{center}
\includegraphics[width=70mm]{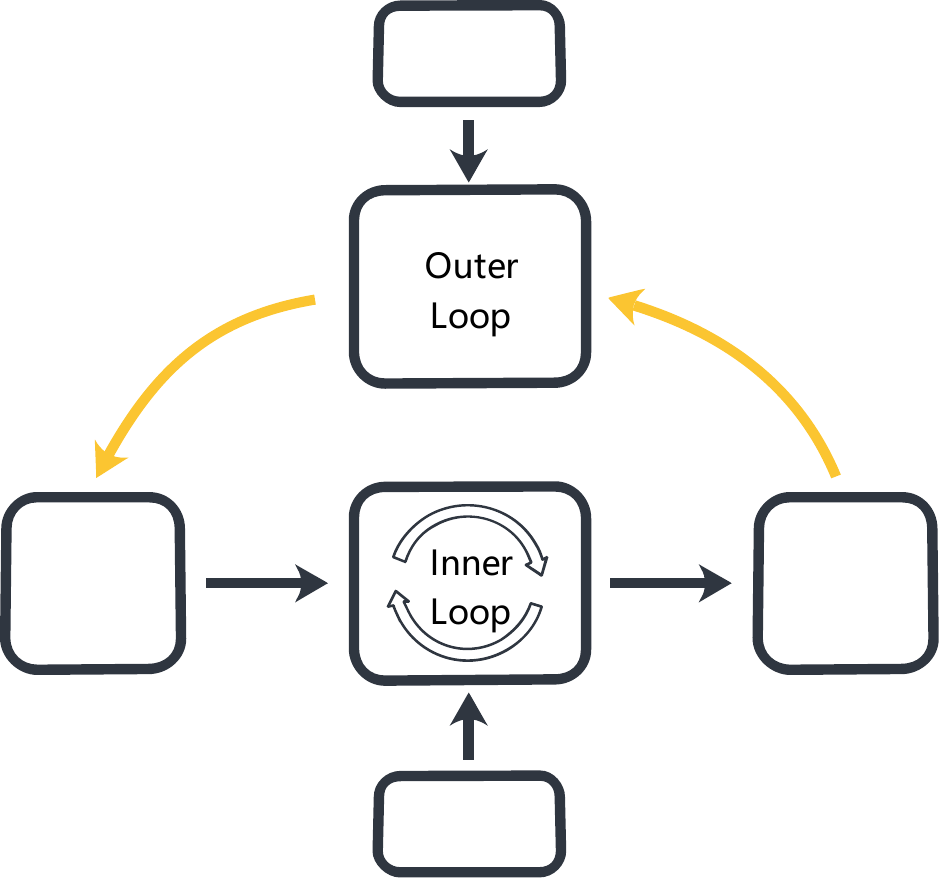}
\put(-188,57){\huge \color{ELECTROMAGNETIC}{$f_\theta$}}
\put(-30,57){\huge \color{ELECTROMAGNETIC}{$f_{\theta'_i}$}}
\put(-114,8){\large \color{ELECTROMAGNETIC}{${\rm x}^{tr}_{ij}$}}
\put(-98,8){\large \color{ELECTROMAGNETIC}{${\rm y}^{tr}_{ij}$}}
\put(-116,172){\color{ELECTROMAGNETIC}{${\rm x}^{test}_{ij}$}}
\put(-100,172){\color{ELECTROMAGNETIC}{${\rm y}^{test}_{ij}$}}
\put(-83,134){\large \color{ELECTROMAGNETIC}{$i$}}
\put(-83,72){\color{ELECTROMAGNETIC}{$j$}}
\put(-230,130){\small \color{RISE-N-SHINE}{$\sum_i\nabla_\theta\mathcal{L}(f_{\psi_i}({\rm x}_{ij}^{test}),{\rm y}_{ij}^{test})$}}
%\put(-63,69){\normalsize \color{ELECTROMAGNETIC}{$\nabla\mathcal{L}_{\psi_i}$}}
\end{center}
   \caption{A typical meta-learning model, where $\theta$ is the parameters updated during the ``outer loop'', $\theta'_i$ represents the parameters obtained by the training on task $i$ during the ``inner loop'', $\{{\rm x}^{tr}_{ij},{\rm y}^{tr}_{ij}\}_j$ and $\{{\rm x}^{test}_{ij},{\rm y}^{test}_{ij}\}_j$ are training examples and testing examples regarding task $i$ respectively.}
\label{MetaGraph}
\end{figure}

Few-shot learning approaches are commonly implemented by meta-learning mechanisms, which enables learn-to-learn. %This subsection draws the basic structure and categories of meta-learning.
There are usually two hierarchies learning processes in meta-learning. The low-level learning process, usually termed as the ``inner loop'', learns to handle general tasks; and the high-level learning process, usually termed as the ``outer loop'', improves the performance of low-level learning process.

Deep learning is often employed in the meta-learning process, although other machine learning methods, such as Bayesian learning~\cite{Lake}, can also be applied. Therefore, most of the meta-learning approaches use gradient descent in the ``outer loop''; hence, the gradient of the ``outer loop'' is termed as meta-gradient. However, the machine learning methods used in the ``inner loop'' are different. According to different machine learning methods meta-learning approaches can be implemented in three categories: 1) memory-based methods, 2) optimization-based methods, and 3) metric-based methods~\cite{Andrei}.

{\bfseries Metric-based methods} can be artificially viewed as using a K-nearest neighbor (KNN) or its variation, to optimise a feature embedding space during the ``outer loop'', minimizing the similarity metrics between instances from same class and maximizing those from different classes, and to predict the labels of testing examples based on the similarity metrics in this embedding space during the ``inner loop''. A number of similarity metrics approaches have been used in meta-learning, such as cosine distance, squared Euclidean distance, or even a relationship learned by a neural network~\cite{Oriol,Flood,Mishra}.

%A number of approaches have been proposed for learning similarity metrics between training and testing examples, which are termed as support and query set~\cite{Koch,Oriol,Jake,Flood,Jinchao,Oreshkin}. {\color{orange}Typical similarity metrics learning adopts the triplet loss in face recognition, which reduces the metric values between in-class samples, and increases those between-classes.} This is often achieved through a {\color{orange}softmax} operation of probability over distances based on a certain metric~\cite{Jake,Oreshkin} while minimizing the negative log-probability. {\color{orange}There are a good collection of distance measures can be used, such as cosine distance, squared Euclidean distance, or even a relationship learned by a neural network}~\cite{Oriol,Flood,Mishra}, although it may not be a strict distance due to the lack of symmetry property.

{\bfseries Optimization-based methods} adopt deep learning methods during the ``inner loop'', and during the ``outer loop'' these approaches learn the hyperparameters of the deep learning methods, such as parameter initialization, learning rate, gradient direction, and et al. Among such methods MAML is the most typical one~\cite{Chelsea}, which tried to learn the parameter initialization. Besides, the methods in \cite{Sachin} and \cite{Fengwei} try to learn the learning rate of the ``inner loop''.

{\bfseries Memory-based methods} remember and search key training examples~\cite{Santoro} or model-related information during the ``inner loop''. The model related information is any information related to the model of ``inner loop'', such as the network weights~\cite{Tsendsuren} or the activation values~\cite{Munkhdalai} of different layers. These methods extend the external memory, and read and write the memory by employing the attention models. %Idea-wise, this is similar to Neural Turing Machines~\cite{Graves}, which imitates operations of computer's read and write operations.

A general meta-learning model is illustrated in Fig.~\ref{MetaGraph}. If a parametric learning method is used during the ``inner loop'', the model will get the parameters $\theta'_i$ by training on the data of task $i$; otherwise, if a nonparametric learning method is used, $\theta'_i$ is just equal to $\{\theta,\{{\rm x}^{tr}_{ij},{\rm y}^{tr}_{ij}\}_j\}$. During the ``inner loop'' $\theta$ is updated to $\theta'_i$ by an algorithm, which is able to keep differentiable between $\theta$ and $\theta'_i$. Last, during the ``outer loop'', $\theta$ is updated by the gradient descent.

%\begin{figure}
%\begin{center}
%\includegraphics[width=50mm]{MetaGraph.pdf}
%\put(-130,67){\Large \color{ELECTROMAGNETIC}{$\theta$}}
%\put(-89,69){\large \color{ELECTROMAGNETIC}{$\psi_i$}}
%\put(-57,118){\large \color{ELECTROMAGNETIC}{${\rm x}^{tr}_{ij}$}}
%\put(-34,88){\large \color{ELECTROMAGNETIC}{${\rm y}^{tr}_{ij}$}}
%\put(-60,19){\large \color{ELECTROMAGNETIC}{${\rm x}^{test}_{ij}$}}
%\put(-37,49){\large \color{ELECTROMAGNETIC}{${\rm y}^{test}_{ij}$}}
%\put(-8,6){\large \color{ELECTROMAGNETIC}{$i$}}
%\put(-16,16){\large \color{ELECTROMAGNETIC}{$j$}}
%\put(-118,82){\normalsize \color{ELECTROMAGNETIC}{$\nabla\mathcal{L}_\theta$}}
%\put(-63,69){\normalsize \color{ELECTROMAGNETIC}{$\nabla\mathcal{L}_{\psi_i}$}}
%\end{center}
%   \caption{{\color{orange}A typical meta-learning model, where $\theta$ is the parameter updated during the ``outer loop'', $\psi_i$ represents the parameters obtained by the training on task $i$ during the ``inner loop'', $\{{\rm x}^{tr}_{ij},{\rm y}^{tr}_{ij}\}_j$ and $\{{\rm x}^{test}_{ij},{\rm y}^{test}_{ij}\}_j$ are training examples and testing examples regarding task $i$ respectively.}}
%\label{MetaGraph}
%\end{figure}

\section{Proposed Meta-Learning Models}\label{Sec_FPMR}

The proposed model is an optimization-based method which controls the gradient descent process during the ``inner loop'' by limiting the model parameters in a low-dimensional latent space. The latent code in the latent space is decoded to the model parameters of a predict model by the non-linear decoder. Several decoders can be chosen based on a different task, so the low-dimensional latent space used in DCN is dependent on the task.

%a predict model with task embedding, named as MHPM. This idea-wise is similar to those of \cite{Andrei,Oreshkin} adapting task embedding in the model to improve the system performance, but the spatial complexity of the proposed work is much lower.

%\ref{sec:TheFPMRAlgorithm}, \ref{Weight_Sharing} and \ref{CMF} introduce the framework and two key structures of {\color{brown}MHPM}, respectively. Its training strategy is detailed in  \ref{Meta-Training_Strategy}. The last two subsections \ref{Ensemble_Learning} {\color{green}and \ref{Dynamic_Routing}} introduce the other improved technologies used in this work.

\subsection{Decoder Choice Network}\label{sec:TheFPMRAlgorithm}
\begin{figure}[t]
\begin{center}
\includegraphics[width=60mm]{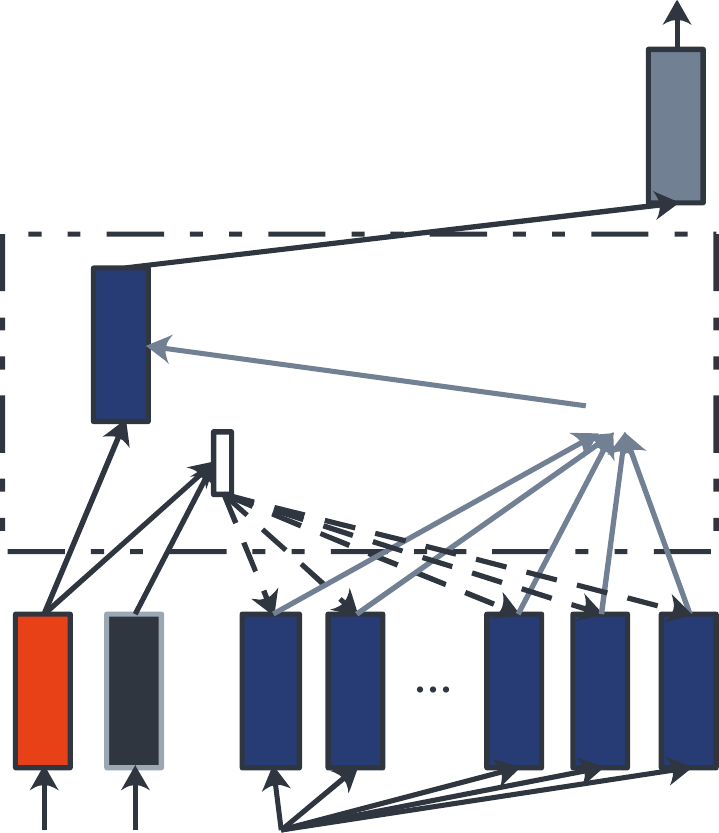}
\put(-106,-10){\large \color{ELECTROMAGNETIC}{${\rm z}$}}
\put(-92,-10){\small \color{ELECTROMAGNETIC}{Latent Code}}
\put(-28,97){\large \color{ELECTROMAGNETIC}{$\hat{\theta}_i$}}
%\put(5,35){\color{ELECTROMAGNETIC}{${\rm R}_\phi$}}
\put(-172,56){\color{ELECTROMAGNETIC}{${\rm x}^{tr}_i$}}
\put(-148,56){\color{ELECTROMAGNETIC}{${\rm y}^{tr}_i$}}
%\put(-108,100){\small \color{ELECTROMAGNETIC}{Membership}}
%\put(-103,90){\small \color{ELECTROMAGNETIC}{Function}}
\put(-111,84){\color{ELECTROMAGNETIC}{$\mathcal{C}$}}
\put(-162,112){\large \color{ELECTROMAGNETIC}{$f_{\theta_i}$}}
\put(-108,114){\small \color{ELECTROMAGNETIC}{Rescale and Shift}}
\put(-76,90){\small \color{ELECTROMAGNETIC}{Integrate}}
\put(-162,146){\small \color{ELECTROMAGNETIC}{DNC}}
\put(-95,170){\small \color{ELECTROMAGNETIC}{Activator Function}}
\put(-99,160){\small \color{ELECTROMAGNETIC}{and other Operations}}
%\put(0,190){\large \color{ELECTROMAGNETIC}{${\rm x}_{t+1}$}}
\put(-110,32){\color{LYNX WHITE}{$d_1$}}
\put(-90,32){\color{LYNX WHITE}{$d_2$}}
\put(-50,32){\color{LYNX WHITE}{$\cdot$}}
\put(-29,32){\color{LYNX WHITE}{$\cdot$}}
\put(-12,32){\color{LYNX WHITE}{$d_S$}}
\end{center}
   \caption{The typical structure of a neural network layer with DCN, where ${\rm x}^{tr}_i=\{{\rm x}_{ij}^{tr}\}_j$ and ${\rm y}^{tr}_i=\{{\rm y}_{ij}^{tr}\}_j$, and the choice network $\mathcal{C}$ is denoted as the white box.}
\label{FuzzyLayer}
\end{figure}
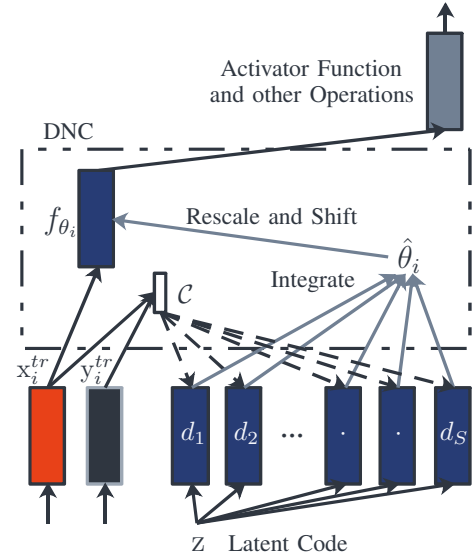

\begin{algorithm}[t]
  \caption{Inner Loop of DCN.}
  \begin{algorithmic}[1]
    \Require
      Choice network $\mathcal{C}$;
      Decoders $\{d_1,d_2,\cdots,d_S\}$;
      Latent code $\rm z$;
      Training examples $\mathcal{D}^{tr}_i=\{{\rm x}^{tr}_{ij},{\rm y}^{tr}_{ij}\}_j$;
      Testing examples $\mathcal{D}^{test}_i=\{{\rm x}^{test}_{ij},{\rm y}^{test}_{ij}\}_j$;
      Learning rate $\alpha$;
      Number of steps $M$;
      Loss function $\mathcal{L}$.

      \State Initialize ${\rm z}'={\rm z}$
      \State $\{c_1,c_2,\cdots,c_S\}\gets\mathcal{C}(\mathcal{D}^{tr}_i)$
      \For {$m=1,\cdots,M$}
        \State $\hat{\theta}_i\gets\sum^S_{s=1}c_s\cdot d_s({\rm z}')$
        \State $\theta_i\gets$ Normalize $\hat{\theta}_i$ by Eq. (\ref{rescale})
        \State $\mathcal{L}^{tr}_i=\sum_j\mathcal{L}(f_{\theta_i}({\rm x}^{tr}_{ij}),{\rm y}^{tr}_{ij})$
        \State ${\rm z}'\gets{\rm z}'-\alpha\nabla_{\rm z'}\mathcal{L}^{tr}_i$
      \EndFor
      \State $\hat{\theta}_i\gets\sum^S_{s=1}c_s\cdot d_s({\rm z}')$
      \State $\theta_i\gets$ Normalize $\hat{\theta}_i$ by Eq. (\ref{rescale})
      \State $\mathcal{L}^{test}_i=\sum_j\mathcal{L}(f_{\theta_i}({\rm x}^{test}_{ij}),{\rm y}^{test}_{ij})$
      \State \Return $\mathcal{L}^{test}_i$
  \end{algorithmic}
\label{InnerLoopDCN}
\end{algorithm}

DCN consists of three parts: choice network $\mathcal{C}$, decoders $\{d_1,d_2,\cdots,d_S\}$ and latent code $\rm z$. In the start of ``inner loop'', the choice network receives the task features and produces the choice of the decoder. In order to make the choice differentiable, here we adopt the idea of Neural Turing Machine (NTMs)~\cite{Graves}. Choice network chooses every decoder with different extends $\{c_1,c_2,\cdots,c_S\}$. The difference with NTMs is that the weights of decoders are not provided by the attention model, but by fuzzy set, and we will detail this in the next subsection.

%{\color{green}The mapping network of each task is built by a linear combination of a set of base mapping networks. The weights of base mapping networks are calculated depending on task by a capsule network~\cite{Sara} and the method of fuzzy set~\cite{dubois1980fuzzy}.} By this way, we build a mapping network depending on the task. {\color{green}After all model parameters mapped from latent code by all base mapping networks and their weights are obtained, they are combined to model parameters of mapping network of this task. This process is depicted in Fig.~\ref{FuzzyLayer}.}

Then, given the weights of decoders, we initialize the latent code of ``inner loop'' latent code ${\rm z}'={\rm z}$ and decode it to the parameters of a neural network model:
\begin{equation}
\label{decode}
\hat{\theta}_i\gets\sum^S_{s=1}c_s\cdot d_s({\rm z}').
\end{equation}
Before parameterizing the neural network model with $\hat{\theta}_i$, in order to prevent vanishing gradient and accelerate convergence $\hat{\theta}_i$ will be normalized by Batch Normalization~\cite{ioffe2015batch}, which is given by:
\begin{equation}
\label{rescale}
\theta_i=\gamma*\frac{\hat{\theta}_i-\mu}{\sqrt{\sigma^2+\epsilon}}+\beta,
\end{equation}
where $\mu$ and $\sigma$ are mean and variance of $\{\hat{\theta}_i\}_i$ respectively, $\gamma$ and $\beta$ are learnable parameters, $\epsilon$ is a positive value close to zero added to the denominator for numerical stability. After obtaining $\theta_i$, the model $f_{\theta_i}$ is builded depending on the task. $f_{\theta_i}$ is used to process each data from training examples with a general feed-forward mapping. The forward process is depicted in Fig.~\ref{FuzzyLayer}.

After getting prediction and loss of all training examples, gradient descent is used to update $\theta$. However, we would not directly update $\theta$, but update the latent code $\rm z$ instead:
\begin{equation}
\label{zup}
{\rm z}'\gets{\rm z}'-\alpha\nabla_{\rm z'}\mathcal{L}^{tr}_i(f_{\theta_i}),
\end{equation}
where $\alpha$ is the learning rate of gradient descent during the ``inner loop'', in order to be simple we rewrite $\sum_j\mathcal{L}(f_{\theta_i}({\rm x}^{tr}_{ij}),{\rm y}^{tr}_{ij})$ as $\mathcal{L}^{tr}_i$. In next step the model parameterized by $\theta_i$ would be used to calculate the loss on training examples, and the process $(\ref{decode})\rightarrow(\ref{rescale})\rightarrow(\ref{zup})\rightarrow(\ref{decode})$  would loop several times before it being evaluated on testing examples. The ``inner loop'' process of DCN is summarised in Algorithm \ref{InnerLoopDCN}.

Finally, the adapted parameters $\theta_i$ are used to calculate the testing loss $\sum_j\mathcal{L}_i(f_{\theta_i}({\rm x}^{test}_{ij}),{\rm y}^{test}_{ij})$, and we write it as $\mathcal{L}^{test}_i$. During the ``outer loop'' the choice network $\mathcal{C}$, the decoders $\{d_1,d_2,\cdots,d_S\}$ and the latent code $\rm z$ are updated to reduce $\mathcal{L}^{test}_i$.

\subsection{Decoders}\label{Weight_Sharing}
%Because network with MHPM is trained end-to-end by meta-gradient, mapping network is trained by the backward gradient from ``outer loop'' loss directly. In this subsection, we detail its structure and analysis.

The main challenge for the implementation of DCN is that if we use the fully connected multi-layer network as decoders it will require too many parameters. The decoders receive the latent code and produce the model parameters of a neural network. Because the dimension of the model parameters is too large, the complexity of decoders become unacceptable. In this subsection, we detail several methods to reduce the complexity of decoders and analyze it.

\subsubsection{Structure of linear transformation}\label{TMRSec}

\begin{figure}[t]
\begin{center}
\includegraphics[width=85mm]{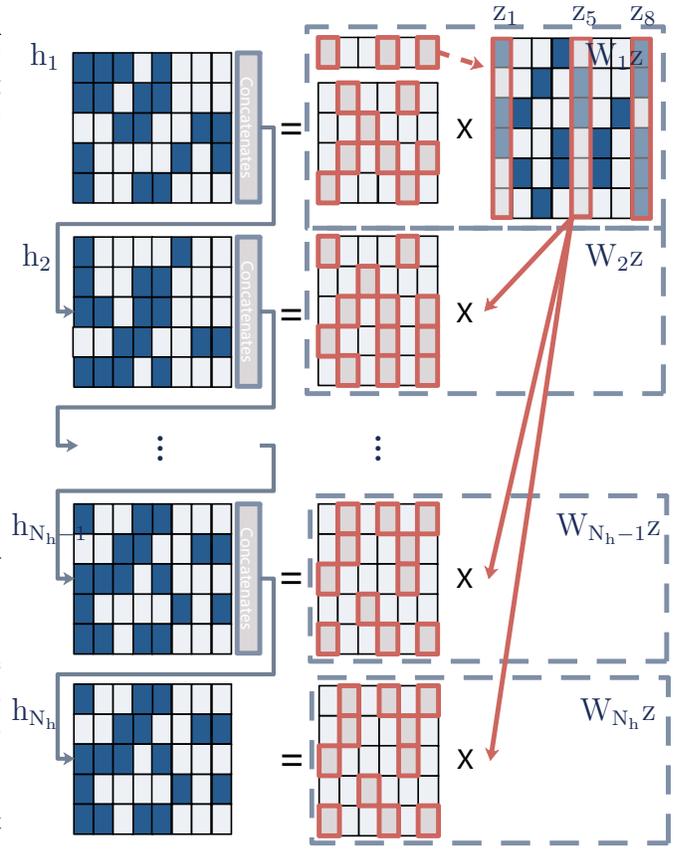}
    \put(-248,296){\large \color{PICO VOID} $\rm {\rm h}_{1}$}
    \put(-251,220){\large \color{PICO VOID} $\rm {\rm h}_{2}$}
    \put(-255,118){\large \color{PICO VOID} $\rm {\rm h}_{N_h-1}$}
    \put(-255,48){\large \color{PICO VOID} $\rm {\rm h}_{N_h}$}
    \put(-38,296){\large \color{PICO VOID} $\rm W_{1}z$}
    \put(-38,220){\large \color{PICO VOID} $\rm W_{2}z$}
    \put(-49,118){\large \color{PICO VOID} $\rm W_{N_h-1}z$}
    \put(-40,48){\large \color{PICO VOID} $\rm W_{N_h}z$}
    \put(-73,313){\large \color{PICO VOID} ${\rm z}_1$}
    \put(-43,313){\large \color{PICO VOID} ${\rm z}_5$}
    \put(-21,313){\large \color{PICO VOID} ${\rm z}_8$}
\end{center}
   \caption{Where 'Concatenates' denotes concatenating $\{\rm h_{1},h_{2},\cdots,h_{N_h}\}$ to $\rm h$. The length of vector in each channel of $\rm h$ is increased by $N_h$ times, which is the number of the weight matrices.}
\label{TimesMapRule}
\end{figure}

Although the main challenge is the high dimension output of decoders, we should also consider the dimension of the latent code $\rm z$ for it can be 10 times smaller than the model parameters at most. The reason is that if the dimension of the latent code is too low it will limit the expression of the meta-learning model.

In order to overcome these two challenges, we divide the input and the weight matrices of a layer in the decoders into groups, and each of the weight matrices is reused in all groups of the input. Here we use the first layer of the decoders as an example to illustrate the idea. The latent code is divided into several groups ${\rm z}=[{\rm z}_1,{\rm z}_2,\cdots,{\rm z}_{N_g}]$, and where $\rm z$ is a matrix and the elements in each column are in the same group. Each element of the output only depends on the latent code in one group. Then each weight matrix of $\{{\rm W}_1,{\rm W}_2,\cdots,{\rm W}_{N_h}\}$ is used to calculate the hidden variable:
\begin{equation}
\begin{aligned}
{\rm h}_n={\rm W}_n {\rm z},\ (n=1,2,\cdots,N_h).
\end{aligned}
\end{equation}

$\{{\rm h}_1,{\rm h}_2,\cdots,{\rm h}_{N_h}\}$ are concatenated to a matrix $\rm h$, and $\rm h$ have already been divided into $N_h$. This process is summarized in Fig.~\ref{TimesMapRule}. We call this structure as group linear transformation (GLT), and we will analyze this structure reduce how much parameters in the decoder network in \ref{Complexity_Analysis}.

%{\color{green}where ${\rm h}^{(k)}=\{{\rm h}_{(k-1)s},{\rm h}_{(k-1)s+1},\cdots,{\rm h}_{ks}\}$, ${\rm h}_i\ (i=1,2,\cdots,Ns)$ is a column vector, and $s$ is the dimension of each input channel. Finally, the output of parallel mapping is ${\rm h}=\{{\rm h}_{1},{\rm h}_{2},\cdots,{\rm h}_{Ns}\}$. This process is more intuitive in Fig.~\ref{TimesMapRule}. Each row in ${\rm h}^{(k)}$ can be viewed as a linear transformation of the valuables in each input channel, which is equal to the operation of a 1d convolutional layer with kernel size 1. Then we increase the number of output variables by N times by using N identical structures.}

%{\color{green}In all experiments we use 2 layers parallel mapping, the second layer outputs $S$ vectors with size $w_1*w_2$.}

\subsubsection{Non-linear}
In hidden layer, we choose ELU ($\alpha=1$)~\cite{clevert2015fast} as non-linear transformation in the decoder network, which is given by:

\begin{equation}
{\rm ELU}(x)=\max(0,x)+\min(0,\alpha*(\exp(x)-1)),
\end{equation}
where $\alpha$ is equal to 1 in all our experiments. In output layer, we use double-thresholding strategy, which has equal thresholds in both positive and negative sides, called ``softshrink''. We use this function based on PyTorch~\cite{Pytorch}. Softshrink is given by:

\begin{equation}
softshrink(x)=\left\{
\begin{aligned}
&x-\lambda,\ \ {\rm if}\ x>\lambda\\
&x+\lambda,\ \ {\rm if}\ x<-\lambda\\
&0,\ \ \ \ \ \ \ {\rm otherwise}
\end{aligned}
\right.,
\end{equation}
where $\lambda$ denotes the threshold. All experiments in this work use $\lambda=0.01$.

\subsubsection{Structure sharing}\label{Multi_Heads}
%As we mention before base mapping network set is denoted by a mapping network with multi heads.
In order to further reduce the complexity of decoders, all decoders network share all low-level layers. Last layer of this neural network has multi heads, the output dimension of each of them is equal to each other, and each head represents a decoder. When the function of each head is related to each other, this kind structure not only reduces the time and spatial complexity of the model but also prevents vanishing gradient, and accelerates convergence greatly. The reason is the shared shallow layers of the network can obtain the gradients from all heads. This is a common method in deep learning, and a number of works have proven its efficiency by empirical evidence, such as \cite{ren2015faster,zhang2016joint,schulman2017proximal}.

We also reuse the choice network and decoders in different layers. However, if the task features are the same for different layers, all layers with DCN would have the same $\{c_1,c_2,\cdots,c_S\}$. We simply use the features of training examples of the previous layer as the task features of the current layer. By this way, each layer chooses its own decoder. In all experiments, we use 2 layers neural network as the decoder, and the second layer outputs are S vectors with the size of the model parameter.

In addition, in some case like miniImageNet, the number of the model parameters is still too large, we resize $\hat{\theta}$ by linear interpolation to enlarge the dimension of $\hat{\theta}$. This method not only reduces the output size of the decoders, but also makes each model parameters related to others, which can be seen as a kind of regularization~\cite{bansal2018can}. Besides, linear interpolation allows DCN to be reused in layers with different dimension model parameters.

\subsubsection{Complexity Analysis}\label{Complexity_Analysis}
We analyze the number of the model parameters in the predict model at first. Assume the number of channels (corresponding to the convolution layer or corresponding to fully connected layer) is $F$. If the number of channels is equal to each other in hidden layers, it would be proportional to $F^2$, and the proportion is $N_lK_s$, where $N_l$ is the number of layers and $K_s$ is kernel size of the convolutional unit. It is obvious that most of CNNs meet this condition. The deviation caused by the first and last layers can be ignored due to the lower magnitude.

We assume that each decoder is a single fully connected network. The input and the output of the decoder network are the latent code and the model parameters, respectively; thus it is sensible to assume ${\rm dim}({\rm z})\propto {\rm dim}(\theta_i)$, which leads to ${\rm dim}({\rm W})\propto F^4N_l^2K_s^2$, where ${\rm W}$ denotes the weight matrix of the decoder network. Since the time and spatial complexity are proportional to the number of the weights in a neural network, both time and spatial complexity are $O(F^4N_l^2K_s^2)$.

Now, we share all decoders amongst the different layers. The time and spatial complexity are reduced to $O(F^4SK_s^2)$. If we the decoders are also shared between the dimension of the kernel, the complexity would reduce to $O(F^4S)$. In general case there is $S\ll N_l^2K_s^2$. Then we replace the fully connected network as the structure proposed in \ref{TMRSec}, GLT.

The number of parameters in 2 layers fully connected network and GLT are $[{\rm dim(z)}+{\rm dim}(\theta_i)S]{\rm dim(h)}$ and $ \frac{\rm dim(h)}{\rm dim(z)}N_g^2+\frac{{\rm dim}(\theta_i)}{\rm dim(h)}SN_h^2$, respectively, where $\rm dim(h)$ and $\rm dim(z)$ are total dimension of the latent code and the hidden variable, respectively. Here $\frac{\rm dim(h)}{\rm dim(z)}$ and $\frac{{\rm dim}(\theta_i)}{\rm dim(h)}$ should be integers, and it is obvious that $1\leq N_g\leq{\rm dim(z)}$ and $1\leq N_h\leq{\rm dim(h)}$. The smaller $N_g$ and $N_h$, the less the number of parameters in GLT. Therefore we can reduce parameters number by reducing the number of the groups. If $N_g=N_h=1$ and ${\rm dim(h)}^2={\rm dim(z)dim}(\theta_i)$, time and spatial complexity are reduced to $O(\frac{F}{\rm dim(z)})$. On the contrary, if $N_g={\rm dim(z)}$ and $N_h={\rm dim(h)}$, GLT is equal to the fully connected network. However, smaller $N_g$ and $N_h$ mean that each element of the model parameters depends on less variables in the latent code. However, in experiments, we find it has enough flexible to fit different tasks even that it is not every element of the model parameters depending on all variables in the latent code.

\subsection{Choice Network}\label{CMF}
Choice network $\mathcal{C}$ receives the task features and produces the choice of the decoder. How to choose task features, using which kind of neural network to process the task features and how to calculate the weights for each decoder should be considered.

On the one hand, the choice network is reused in different layers. In order to choose different decoders in different layers, the choice network receives the input features of all training examples in the current layer as the task features of this layer. By this way, each layer chooses its own decoder. On the other hand, the choice network is also reused in the different dimension of the convolutional network kernel. The task features should be organized based on the dimension of kernels, which is detailed in Appendix \ref{Taskfeatkernel}.

After obtaining the task features, capsule net~\cite{Sara} is adopted to process the task features. We use 1 capsule layer in all our experiments. The task features which are input to the capsule layer are divided into several capsules, and each capsule only involves 1 variable. The output variables of the capsule layer are all in one capsule. Since \cite{Sara} has detailed the process of dynamic routing, more details are illustrated in Appendix \ref{State_variables}.

Finally, fuzzy set is used to calculate the weight of each decoders. The output variables of the capsule layer are in $[-1,1]$, which are denoted as $\{v_1,v_2,\cdots,v_{N_f}\}$. We transform the output variables to $[0,1]$ to obtain the state variables, $\gamma_n=(v_n+1)/2,\ n=1,2,\cdots,N_f$.
%\subsubsection{Firing strength}\label{Firing_strength}
Each state variable is represented into two fuzzy sets, so there is $S=2^{N_f}$. The relationship of two firing strengths is $\mu_A(x)=1-\mu_B(x)$, and the value of $\mu_A$ is calculated by,
\begin{equation}
\mu_A(x)=\left\{
\begin{aligned}
&1,\ \ \ \ \ \ \ x\leq0\\
&1-x,\ \ 0<x\leq1\\
&0,\ \ \ \ \ \ \ x>1
\end{aligned}
\right..
\end{equation}
The firing strengths of two fuzzy set in $\gamma_n$ are $\mu_A(\gamma_n)$ and $1-\mu_A(\gamma_n)$. The weight of each decoder is calculated by:
\begin{equation}
\label{weig}
c_s=\frac{\prod_n\mu_A(\gamma_n)^{a^n_s}\mu_B(\gamma_n)^{1-a^n_s}} {\sum^S_{l=1}\prod_n\mu_A(\gamma_n)^{a^n_l}\mu_B(\gamma_n)^{1-a^n_l}},
\end{equation}
where $a^n_s\in\{0,1\}$ denotes the $s^{th}$ variable has which fuzzy set. Due to $\sum^S_{l=1}\prod_n\mu_A(\gamma_n)^{a^n_l}\mu_B(\gamma_n)^{1-a^n_l}\equiv1$, Eq. (\ref{weig}) can be written as:
\begin{equation}
\label{weig}
c_s=\prod_n\mu_A(\gamma_n)^{a^n_s}\mu_B(\gamma_n)^{1-a^n_s}.
\end{equation}

For the complexity of the choice network: First, the complexity of capsule net is $O(N_{\rm x}N_f)$, where $N_{\rm x}$ is the dimension of data features. On the one hand, $N_{\rm x}$ would not larger than the dimension of the original feature. The reason is that we will use the feature embedding to process data in order to get low-dimensional, highly abstract features (see \ref{Meta-Training_Strategy}). On the other hand,  $N_f=\log{S}$ and the largest $S$ is equal to 16 in all experiments, $N_f$ is very small; therefore, the size of capsule net is a fairly small model relative to the decoders. Second, the complexity of calculating firing strength is $O(S\log{S})$. This complexity is also can be ignored relative to the complexity of decoders. As a result, the time and spatial complexity of DCN are similar to those of the decoders with little effect from the choice network.

\subsection{Meta-Training Strategy}\label{Meta-Training_Strategy}

\begin{figure*}
\begin{center}
\includegraphics[width=150mm]{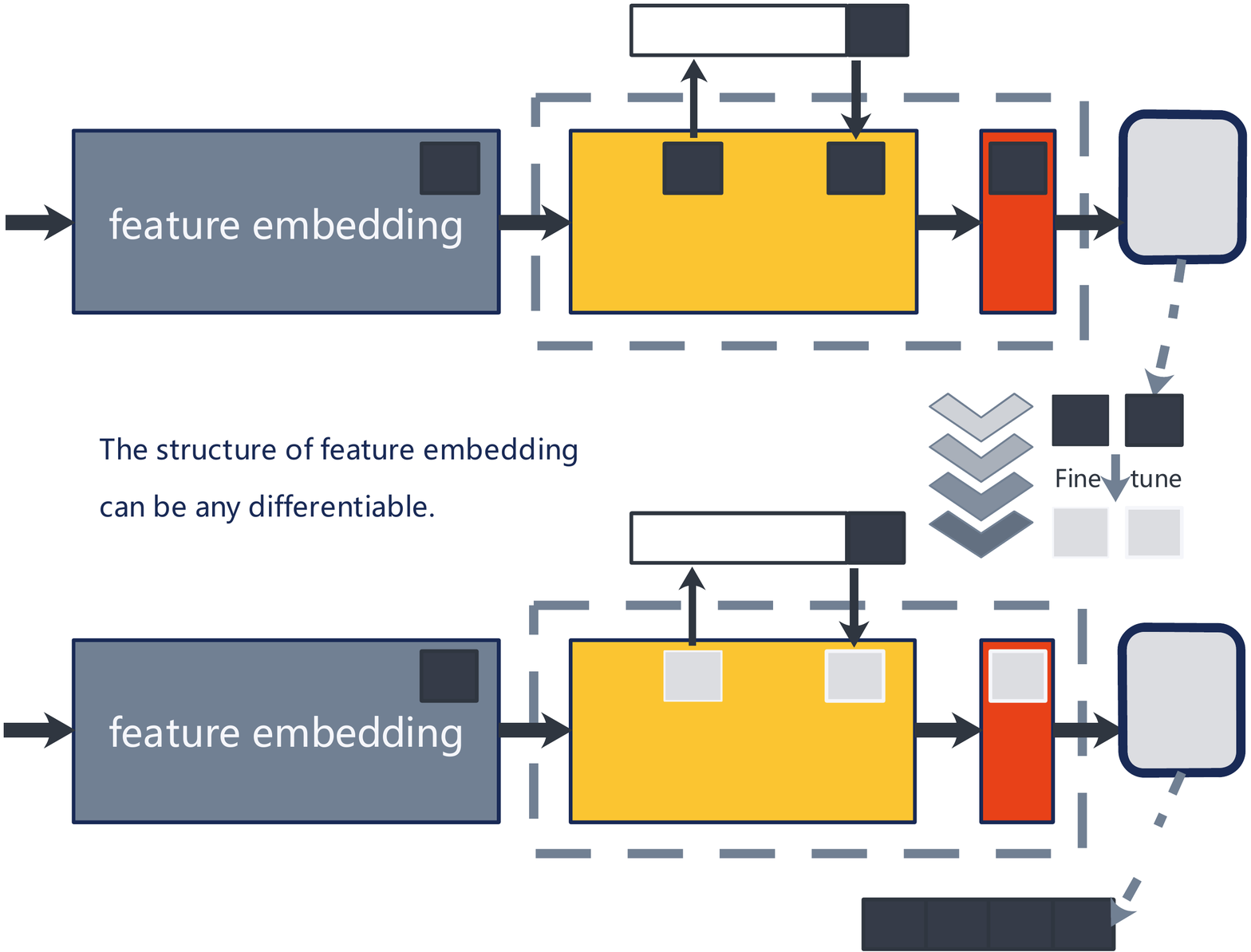}
%    \put(-445,228){\large \color{ELECTROMAGNETIC} $\mathcal{D}^{tr}_{feat}$}
%    \put(-56,250){\large \color{ELECTROMAGNETIC} $\mathcal{D}^{tr}_{label}$}
%    \put(-56,228){\large \color{ELECTROMAGNETIC} $\mathcal{P}^{tr}_{label}$}
%    \put(-445,67){\large \color{ELECTROMAGNETIC} $\mathcal{D}^{test}_{feat}$}
%    \put(-56,88){\large \color{ELECTROMAGNETIC} $\mathcal{D}^{test}_{label}$}
%    \put(-56,65){\large \color{ELECTROMAGNETIC} $\mathcal{P}^{test}_{label}$}
    \put(-435,228){\large \color{ELECTROMAGNETIC} $\mathcal{X}^{tr}$}
    \put(-50,248){\large \color{ELECTROMAGNETIC} $\mathcal{Y}^{tr}$}
    \put(-50,226){\large \color{ELECTROMAGNETIC} $\mathcal{P}^{tr}$}
    \put(-443,67){\large \color{ELECTROMAGNETIC} $\mathcal{X}^{test}$}
    \put(-55,86){\large \color{ELECTROMAGNETIC} $\mathcal{Y}^{test}$}
    \put(-55,63){\large \color{ELECTROMAGNETIC} $\mathcal{P}^{test}$}
    \put(-66,196){\large \color{ELECTROMAGNETIC} $\nabla_{{\rm z},\theta_{\rm FC}}\mathcal{L}^{tr}$}
    \put(-68,28){\large \color{ELECTROMAGNETIC} $\nabla_{\Theta}\mathcal{L}^{test}$}
    \put(-215,128){\small \color{ELECTROMAGNETIC} $\mathcal{C}$, \{$d_1$,$d_2$,...,$d_S$\}}
    \put(-145,128){\small \color{LYNX WHITE} $\theta_{cd}$}
    \put(-215,289){\small \color{ELECTROMAGNETIC} $\mathcal{C}$, \{$d_1$,$d_2$,...,$d_S$\}}
    \put(-145,288){\small \color{LYNX WHITE} $\theta_{cd}$}
    \put(-280,245){\small \color{LYNX WHITE} $\theta_{fe}$}
    \put(-200,245){\small \color{LYNX WHITE} $\rm z$}
    \put(-149,244){\small \color{LYNX WHITE} $\theta$}
    \put(-102,245){\small \color{LYNX WHITE} $\theta_{\rm FC}$}
    \put(-280,84){\small \color{LYNX WHITE} $\theta_{fe}$}
    \put(-200,84){\small \color{ELECTROMAGNETIC} $\rm z'$}
    \put(-149,84){\small \color{ELECTROMAGNETIC} $\theta_i$}
    \put(-102,84){\small \color{ELECTROMAGNETIC} $\theta'_{\rm FC}$}
%    \put(-188,226){\Large \color{LYNX WHITE} ${\rm H}_{\theta}$}
%    \put(-188,65){\Large \color{LYNX WHITE} ${\rm H}_{\theta'}$}
    \put(-188,226){\Large \color{LYNX WHITE} $f_{\theta}$}
    \put(-188,65){\Large \color{LYNX WHITE} $f_{\theta_i}$}
    \put(-106,226){\Large \color{LYNX WHITE} ${\rm FC}$}
    \put(-106,65){\Large \color{LYNX WHITE} ${\rm FC}$}
    \put(-78,165){\small \color{LYNX WHITE} $\rm z$}
    \put(-59,165){\small \color{LYNX WHITE} $\theta_{\rm FC}$}
    \put(-78,130){\small \color{ELECTROMAGNETIC} $\rm z'$}
    \put(-59,130){\small \color{ELECTROMAGNETIC} $\theta'_{\rm FC}$}
    \put(-139,6){\small \color{LYNX WHITE} $\theta_{fe}$}
    \put(-116,6){\small \color{LYNX WHITE} $\rm z$}
    \put(-99,6){\small \color{LYNX WHITE} $\theta_{cd}$}
    \put(-81,6){\small \color{LYNX WHITE} $\theta_{\rm FC}$}
    \put(-158,5){\large \color{ELECTROMAGNETIC} $\Theta$}
\end{center}
   \caption{A typical training process, where $\Theta=\{\theta_{fe},{\rm z},\theta_{cd},\theta_{\rm FC}\}$, and $\theta_d$ is the parameters of choice network, decoders; $\rm FC$ denotes the last layer of model; $\mathcal{X}^{tr}=\{x^{tr}_{ij}\}_j$ and $\mathcal{Y}^{tr}=\{y^{tr}_{ij}\}_j$ are the features and the labels of training examples, $\mathcal{X}^{test}=\{x^{test}_{ij}\}_j$ and $\mathcal{Y}^{test}=\{y^{test}_{ij}\}_j$ are those of testing examples, and $\mathcal{P}^{tr}$ and $\mathcal{P}^{test}$ are the model prediction for the training and testing samples. The $\rm z$, is the latent code, is transformed into the model parameters $\theta$ by the decoders network. During inner training step get the training loss $\mathcal{L}_i^{tr}$, and fine tune $(\rm z, \theta_{FC})$ to $(\rm z',\theta'_{FC})$ by $\nabla_{{\rm z},\theta'_{FC}}\mathcal{L}_i^{tr}$ during the ``inner loop''. Then the choice network and the decoders receives $\rm z'$ and product $\theta_i$. Last the testing loss $\mathcal{L}_i^{test}$ was calculated to update $\Theta$ during the ``outer loop''. }
\label{TrainingProcess}
\end{figure*}

\subsubsection{Feature embedding}
It is necessary to use a much deeper network, such as resnet~\cite{He} or dense net~\cite{Gao}, for higher classification accuracy on a dataset with complicated image content such as miniImagenet. However, it requires too a large amount of GPU memory to learn all the parameters of a very deep network by optimization-based methods or memory-based methods. Besides, some metric-based methods would be instability in this case~\cite{Oreshkin}. Therefore, the co-training or supervised pre-training methods have been proposed to ease such situation recently. For examples, the task-dependent adaptive metric (TADAM)~\cite{Oreshkin} trains resnet with auxiliary co-training of 64 classifications~\cite{Oreshkin} in order to improve convergence; Latent embedding optimization (LEO)~\cite{Andrei} used the whole classes from training and validation set to do 80 classification pre-training in order to get a feature embedding, and the similar approach was reported in \cite{Siyuan,Bauer,Spyros}.

In contrast, the proposed method is trained end-to-end. The training process of DCN is summarised in Fig.~\ref{TrainingProcess}. The feature embedding in Fig.~\ref{TrainingProcess} is implemented using the standard CNN or much deep network, such as resnet~\cite{He} or dense net~\cite{Gao}. As Fig.~\ref{TrainingProcess} shows, the parameters of the feature embedding $\theta_{fe}$ is consistent during the ``inner loop'', and updated during the ``outer loop''. Due to the fine-tuning in the last few layers is differentiable, The feature embedding can be updated by gradient backward from the last few layers directly. This method make meta-training a very deep network become simpler and efficiency.

\subsubsection{Fine-tuning}
DCN is an optimization-based method and thus it implements gradient descent in the ``inner loop''. The latent parameters are updated by the method which we explain in \ref{sec:TheFPMRAlgorithm}. The other parameters such as $\theta_{\rm FC}$ in the last fully connected layer (FC), showed in Fig.~\ref{TrainingProcess}, are updated by gradient descent like MAML~\cite{Chelsea}.

The hyperparameters which is adopted to control the ``inner loop'' like the learning rate $\alpha$, the choice network $\mathcal{C}$ and the decoders $\{d_1,d_2,\cdots,d_s\}$ are consistent during the ``inner loop'', and updated during the ``outer loop'' by stochastic gradient descent(SGD).

\begin{algorithm}[t]
  \caption{Outer Loop of DCN.}
  \begin{algorithmic}[1]
    \Require
      Parameters of choice network and decoders $\theta_{cd}$;
      Learning rate of the ``inner loop'' $\alpha$;
      Number of steps of the ``inner loop'' $M$;
      Loss function $\mathcal{L}$;
      Parameters of FC layer and latent code $\theta_{\rm FC},{\rm z}$;
      Distribution over tasks $p(\mathcal{T})$;
      Parameters of the feature embedding $\theta_{fe}$;
      Learning rate of the ``outer loop'' $\beta$;

    \State Initialise $\Theta=\{\theta_{fe},{\rm z},\theta_{cd},\theta_{\rm FC}\}$
    \While{$\Theta$ has not converged}
      \State Sample batch of tasks $\{\mathcal{D}^{tr}_i,\mathcal{D}^{test}_i\}_i\sim p(\mathcal{T})$
      \ForAll{$\{\mathcal{D}^{tr}_i,\mathcal{D}^{test}_i\}$} \verb|\|\verb|\| Inner Loop
        \State Use the feature embedding to extract features:
        \begin{equation}\nonumber
        \hat{\mathcal{D}}^{tr}_i=\{\hat{\rm x}^{tr}_{ij},{\rm y}^{tr}_{ij}\}_j,\ \hat{\mathcal{D}}^{test}_i=\{\hat{\rm x}^{test}_{ij},{\rm y}^{test}_{ij}\}_j
        \end{equation}

        \State Initialize ${\rm z}'={\rm z},\ \theta_{\rm FC}'=\theta_{\rm FC}$
        \State $\{c_1,c_2,\cdots,c_S\}\gets\mathcal{C}(\mathcal{D}^{tr}_i)$ with $\theta_{cd}$
        \For {$m=1,\cdots,M$}
          \State $\hat{\theta}_i\gets\sum^S_{s=1}c_s\cdot d_s({\rm z}')$
          \State $\theta_i\gets$ Normalize $\hat{\theta}_i$ by Eq. (\ref{rescale})
          \State $\hat{\rm x}^{tr}_{ij}\gets f_{\theta_i}(\hat{\rm x}^{tr}_{ij})$
          \State $\mathcal{L}^{tr}_i=\sum_j\mathcal{L}(f_{\theta_{\rm FC}}(\hat{\rm x}^{tr}_{ij}),{\rm y}^{tr}_{ij})$
          \State ${\rm z}'\gets{\rm z}'-\alpha\nabla_{{\rm z}',\theta'_{\rm FC}}\mathcal{L}^{tr}_i$
        \EndFor
        \State $\hat{\theta}_i\gets\sum^S_{s=1}c_s\cdot d_s({\rm z}')$
        \State $\theta_i\gets$ Normalize $\hat{\theta}_i$ by Eq. (\ref{rescale})
        \State $\hat{\rm x}^{test}_{ij}\gets f_{\theta_i}(\hat{\rm x}^{test}_{ij})$
        \State $\mathcal{L}^{test}_i=\sum_j\mathcal{L}(f_{\theta'_{\rm FC}}(\hat{\rm x}^{test}_{ij}),{\rm y}^{test}_{ij})$
      \EndFor

      \State $\Theta\gets\Theta-\beta\nabla_\Theta\sum_i\mathcal{L}^{test}_i$
    \EndWhile
  \end{algorithmic}
\label{OuterLoopDCN}
\end{algorithm}

The ``outer loop'' process of DCN is summarised in Algorithm~\ref{OuterLoopDCN}. Due to $\theta_{\rm FC}$ is also required to updated during the ``inner loop'', it would be a little different from Algorithm~\ref{InnerLoopDCN}. After model initialization, the approach randomly samples a batch of tasks $\{\mathcal{D}^{tr}_i,\mathcal{D}^{test}_i\}_i$ from task distribution $\mathcal{T}$ and extract features of all examples in $\{\mathcal{D}^{tr}_i,\mathcal{D}^{test}_i\}_i$ by the feature embedding.

Then forward propagation of training examples is performed within each task and the backward gradient is used to update the parameters $\theta_{\rm FC}$ and $\rm z$ to get $\theta'_{\rm FC}$ and $\rm z'$ (detailed in the ``inner inner''). After training within each task, a unique model is generated for each task and this model is tested on the testing examples of each task. Finally, the loss of all tasks based on the testing examples is used to update $\Theta$ during the ``outer loop''.

\subsection{Ensemble Learning}\label{Ensemble_Learning}
The standard training protocol used in most of the previous few-shot learning problems is decreasing learning rates during training and choosing a model by the validation set. However, a recent study found a training protocol with snapshot ensemble~\cite{Huang} is more suitable for the model training on some datasets like miniImageNet. This is because it makes better use of the model obtained in a single training process.

This is because when a model is trained based on miniImageNet, there are not the same classes between training and validation set, and the number of classes is too small in training set, there is larger generalization gap between training and validation loss. When the training loss reduces quickly, the validation loss barely changes. In this case, picking models in a different number of iteration can obtain the models with similar validation losses and completely different training losses. It means that there are several which have similar generalization performance, but they are very different from each other, so the diversity and quality of models ensemble are fully guaranteed.

In this work, instead of using the best model in the training process, all models with strong performance sample from the training process are used. This approach has been adopted in support nets~\cite{Jinchao}, and they have proved its effectiveness by experiments. Following \cite{Jinchao}, the proposed work chooses the top $n$ models with the best performance on the validation set as an ensemble model instead of a single model.

We take the average outputs of all models involved in the ensemble model. The models are selected at certain iteration intervals, and sort based on accuracy. Those models are added to the ensemble model in order of accuracy on the validation set, and if one model improves the performance of the ensemble model, it is retained, otherwise dropped. The model with higher validation accuracy is tested for ensemble earlier.

The ensemble learning is integrated into the proposed optimization-based method DCN, and we call DCN with ensemble learning as DCN-E. This combination utilizes all the models obtained in a single training process, and the results show that this method improves the generation performance of the model greatly.

\subsection{Compare DCN with LEO}
%To begin with, the motivations of MHPM and LEO~\cite{Andrei} are different, though both of them are developed from MAML~\cite{Chelsea}.

%{\color{red}The method most similar to the proposed model is LEO~\cite{Andrei}, which encodes the latent code depending on the task and decodes the latent code to the model parameters by decoder. MHPM is opposite to LEO using the same latent code in all tasks, but decode network is built depending on the task, and we call the decode network as mapping network in this work.}

The method which is most similar to the proposed is LEO~\cite{Andrei}, which encodes the latent code depending on the task and decodes the latent code to the model parameters by the decoder. Both LEO~\cite{Andrei} and DCN perform gradient descent within a low-dimensional latent space during the ``inner loop''. However, there are several differences between LEO~\cite{Andrei} and DCN.

To begin with, LEO~\cite{Andrei} and DCN build task dependence on different places. Both of LEO~\cite{Andrei} and DCN try to learn the latent code and the decoder which decodes the latent code to the model parameters. The difference between LEO~\cite{Andrei} and DCN is that LEO~\cite{Andrei} try to build task dependence on the latent code, while DCN try to build task dependence on the decoder. Besides, the decoder of LEO~\cite{Andrei} and the latent code of DCN are constant for different tasks.

In addition, LEO~\cite{Andrei} and DCN build task dependence by a different way. LEO~\cite{Andrei} uses a network to convert the task features to the latent code. DCN chooses the decoder from the set of decoders according to the task features.

Secondly, the model parameters which LEO~\cite{Andrei} and DCN try to generate are different. LEO~\cite{Andrei} generates the model parameter of the last layer, while DCN generates that of the hidden layer. Higher time and space complexity are required for generating the model parameter of the hidden layer. This is because the model parameter of the last layer can be generated by classes within a task separately, and the features of each class are used for the model parameter of this class. However, the model parameter of the hidden layer cannot be separated by classes.

For this reason, the output size of decoder increases from $h_s$ to $h_s^2$, and the parameter size of decoder increases from $h_s^2$ to $h_s^4$. This is the problem we analyzed in Section \ref{Complexity_Analysis}. Therefore, LEO~\cite{Andrei} only uses a linear decoder. However, since DCN can reduce the parameter size of decode to an acceptable scale, DCN can use multi-layers non-linear decoders.

\section{Experimentation}\label{Sec_Experimentation}
A comparative study in reference to the state-of-the-art approaches is reported in this section for model evaluation, and particularly to address these three questions: (1) Can optimization control of DCN learn useful information for meta-learning? (2) Can the proposed training strategy train a much deep network without co- or pre-training? (3) Can snapshot ensemble~\cite{Huang} improves the performance of meta-learning? In this work, all experiments were run on Pytorch~\cite{Pytorch}.

Regression and classification tasks are used in the experiments. The regression task is a simple sinusoid curve fitting. The results are compared with MAML's~\cite{Chelsea}. The classification tasks are based on the Omniglot~\cite{Brenden} and miniImageNet dataset~\cite{Oriol} which are common benchmark few-shot learning tasks. The code is available on \url{www.github.com/AceChuse/DCN}.

\subsection{Data Description}
The {\rm \bfseries Sinusoid} curve fitting task has been used in the work of \cite{Chelsea}. The examples of each task are sampled from the input and output of a sine wine, where the amplitudes and phases of sine wine from $p(\mathcal{T})$ are different. The amplitudes and phases random are sampled from the uniform distribution from 0.1 to 5.0 and 0 to $\pi$, respectively. Both training and testing features are sampled uniformly from $[-5.0,5.0]$. The mean-squared error between the output of the network and corresponding sine function value is used for both evaluation and loss function.

The {\rm \bfseries Omniglot} consists of samples from 50 international languages, each character has 20 instances, there are a total of 1,623 characters. The 20 instances of each character ward are written by a different person. Following the way of \cite{Oriol}, the Omniglot dataset was divided into 1,200 and 423 characters for training and evaluation respectively. All images were resized to $28\times28$, and samples were augmented by being rotated 90, 180, 270 degrees. The model was evaluated on 1-shot and 5-shot, 5-way and 20-way tasks. Each task contains the same number of shot and query.

The {\rm \bfseries miniImageNet} consists of 100 classes each of which involves 600 natural images. We resize all images into $84\times84$ in order to guarantee a fair comparison to prior work. The miniImageNet dataset was sampled from ILSVRC-12 dataset~\cite{Russakovsky}. This dataset was first proposed in \cite{Oriol} by Vinyals et al. Following the split of the miniImageNet proposed by Ravi and Larochelle~\cite{Sachin} and most previous work, the dataset was split to 64, 16 and 20 class for training, validation, and testing, respectively.

%\subsection{Experiment Configuration}
\subsubsection{Sinusoid Curve Fitting}
Three hidden layers of size $[40,40,35]$ were used with ReLU nonlinearities, two layers in the middle are layer with DCN. This is different from ~\cite{Chelsea}, because if two hidden were used in the model, there would be only one layers with DCN. In this case, it is hard to reflect the advantages of DCN, since it needs to choose the decoders which are shared between layers. Except for the hidden layers with DCN, the other layers are the standard fully connected layer trained by MAML. There is not the feature embedding which does not fine-tune in sinusoid curve fitting experiment.

During training process, two step updates were applied with the ``inner loop'' fixed learning rate $\alpha=0.01$, and fixed the ``outer loop'' learning rate $\beta=10^{-3}$. Both of two models are trained for 60,000 iterations by Adam with AMSGrad~\cite{Duchi}. We did not update the inner learning and weight decay rate in sinusoid curve fitting experiment. During testing, we use 10, 20 and 30 steps update for 5-shot, 10-shot, and 20-shot respectively. The common loss function, mean-squared loss, was used for sinusoid curve fitting, the form of the loss is given by,

\begin{equation}
\mathcal{L}^{test}_i=\sum_j\|f_\theta({\rm x}^{test}_{ij})-{\rm y}^{test}_{ij}\|^2_2,
\end{equation}
where ${\rm x}^{test}_{ij}$ and ${\rm y}^{test}_{ij}$ are the input and output sampled from each sinusoid curve, $f_\theta$ denotes a model with parameters $\theta$.

Since the sizes of hidden layers are not equal to each other, we use linear interpolation to resize the output of the decoders. Table~\ref{SineResult} shows the results of both MAML and proposed method. There are 3116 parameters in MAML and 2,020 parameters in DCN, these contain all learnable parameters. Their structure builded in the ``inner loop'' used to forward is same. Due to the unlimited number of samples, using the same scale model to ensure fairness is required. 600 mini-batch of tasks are randomly sampled to evaluate, each batch has 25 tasks.

\renewcommand\arraystretch{1.}
\begin{table}[t]
\caption{\upshape The mean-squared error of sinusoid curve fitting. The 95\% confidence intervals over tasks is showed after $\pm$.}
\begin{center}
\resizebox{1.\hsize}{!}{
\begin{tabular}{c|c|c|c}
\hline
Models & 5-shot & 10-shot & 20-shot \\
\hline
MAML & 0.1564$\pm$0.0052 & 0.0360$\pm$0.0011 & 0.0055$\pm$0.00014 \\
\hline
DCN & 0.0176$\pm$0.0011 & 0.0051$\pm$0.0001 & 0.0028$\pm$0.00005 \\
\hline
\end{tabular}
}
\end{center}
\label{SineResult}
\end{table}

\subsubsection{Omniglot}
The layers with DCN were applied to replace two convolution layers before fully connected layer in a standard 4-layer embedding CNNs which proposed in \cite{Oriol}, and mean-pooling used to replace max-pooling, this model is denoted by DCN4. DCN6 denote the model with four convolution layers and two layers with DCN before fully connected layer. The layers before the layers with DCN is the feature embedding which is not fine-tuned, and their parameters are updated by gradient descent during the ``outer loop''. The fully connected layer with size 64 and softmax was used to calculate the classification probability. As in the case of sinusoid curve fitting, our model has fewer parameters than a standard 4-layer convolution embedding with a fully connected layer and softmax. In the 5-way task, there are 112,005 parameters in a standard 4-layer convolution embedding and 76,553 parameters in DCN4, these contain all learnable parameters, of course. Models were trained for 60K iterations, and the initial learning rate is $10^{-3}$, and decays by 0.5 for every $\rm10K$ episode. Inner learning and weight decay rate were updated in Omniglot experiment, and their learning rate is 0.1 times of other parameters. The common loss function, cross-entropy loss, was used for classification, the loss takes the form:

\begin{equation}
\mathcal{L}^{test}_i=\sum_j\sum_k{\rm y}^{test}_{ij,k}\log f_{\theta,k}({\rm x}^{test}_{ij})
\end{equation}
where ${\rm x}^{test}_{ij}$ is the features of a testing example, $\{{\rm y}^{test}_{ij,k}\}_k$ is the labels of a testing example, and $f_\theta$ denotes a model with parameters $\theta$.

\renewcommand\arraystretch{1.5}
\begin{table*}
\small
\caption{\upshape The results of few-shot classification on Omniglot dataset, which are average accuracy over 1,800 tasks generated from testing set. After $\pm$ is 95\% confidence intervals over testing tasks. The higher accuracies of DCN and LEO with same structure are highlighted.}
\begin{center}
\setlength{\tabcolsep}{5mm}{
\begin{tabular}{lcccc}
\toprule[1pt]
\textbf{Model} & \multicolumn{2}{c}{\textbf{5-way Acc}} & \multicolumn{2}{c}{\textbf{20-way Acc}} \\
     & 1-shot & 5-shot & 1-shot &5-shot \\
\hline
LEO4\cite{Mishra} & 99.433$\pm$0.074\% & 99.727$\pm$0.011\% & 98.358$\pm$0.030\% & 99.178$\pm$0.005\%\\
LEO4-E\cite{Mishra} & 99.444$\pm$0.075\% & 99.727$\pm$0.011\% & 98.361$\pm$0.030\% & 99.184$\pm$0.005\%\\
LEO6\cite{Mishra} & 99.422$\pm$0.073\% & 99.751$\pm$0.012\% & 98.692$\pm$0.026\% & 99.355$\pm$0.005\%\\
LEO6-E\cite{Mishra} & 99.478$\pm$0.070\% & 99.771$\pm$0.010\% & 98.736$\pm$0.026\% & 99.370$\pm$0.004\%\\
\hline
DCN4 & \textbf{99.800$\pm$0.050\%} & \textbf{99.891$\pm$0.008\%} & \textbf{98.825$\pm$0.025\%} & \textbf{99.505$\pm$0.004\%}\\
DCN4-E & \textbf{99.833$\pm$0.042\%} & \textbf{99.909$\pm$0.007\%} & \textbf{98.842$\pm$0.025\%} & \textbf{99.522$\pm$0.004\%}\\
%MHPM4-D & 99.733$\pm$0.049\% & 99.860$\pm$0.009\% & 99.139$\pm$0.022\% & 99.564$\pm$0.004\% \\
%MHPM4-DE & 99.711$\pm$0.051\% & 99.880$\pm$0.009\% & \textbf{99.206$\pm$0.021\%} & 99.610$\pm$0.003\% \\
DCN6 & \textbf{99.856$\pm$0.040\%} & \textbf{99.924$\pm$0.007\%} & \textbf{99.183$\pm$0.021\%} & \textbf{99.593$\pm$0.004\%}\\
DCN6-E & \textbf{99.922$\pm$0.032\%} & \textbf{99.924$\pm$0.007\%} & \textbf{99.108$\pm$0.022\%} & \textbf{99.633$\pm$0.003\%}\\
\bottomrule[1pt]
\end{tabular}
}
\end{center}
\label{OmniglotLEO}
\end{table*}

Since DCN is similar to LEO~\cite{Andrei}, we do the comparative experiments and the results are showed in Table~\ref{OmniglotLEO}. LEO is used to replace DCN in DCN4, DCN4-E, DCN6, and DCN6-E. The coefficients of entropy penalty and stopgrad penalty are set as 0.1 and 1e-8, which have a similar scale with those obtained by random search in \cite{Andrei}, and we have not used orthogonality penalty because it requires too much GPU memory. Since we explain before that LEO requires too many parameters which make it unfair to compare with other model. In 5-way task, there are 5,125,525 parameters in LEO4, and the other three models are also much larger than the model of DCN. However, DCN has better performance in most case. Since DCN4 has the same size as the model used in the prior works, we compare it with other methods in Table~\ref{OmniglotResult}.

%When we proceed experiments on MHPM-D, we only replace the mean-pooling after last layer with MHPM. We calculate 5 features from each class. Following \cite{Sara}, margin loss was used for classification, the loss takes the form:
%
%\begin{equation}
%\begin{aligned}
%\mathcal{L}_{\mathcal{T}_i}(f_\phi)=&\sum_{{\rm x}^{(j)},{\rm y}^{(j)}\sim\mathcal{T}_i}\sum_k\big[{\rm y}^{(j)}_k\max(0,m^+-\|f_{\phi,k}({\rm x}^{(j)})\|)^2\\
%&+\lambda(1-{\rm y}^{(j)}_k)\max(0,\|f_{\phi,k}({\rm x}^{(j)})\|-m^-)^2\big],
%\end{aligned}
%\end{equation}
%where $m^+=0.9$ and $m^-=0.1$, the down-weighting $\lambda=0.5$, these are same as \cite{Sara}. The other part is same as MHPM. Both of our models were trained end to end using Adam with AMSGrad~\cite{Duchi}, and keeping 1 step update in training and testing, and  each batch has 32 and 16 tasks in 5-way and 20-way task respectively.

\renewcommand\arraystretch{1.5}
\begin{table*}
\small
\caption{\upshape The results of few-shot classification on Omniglot dataset, which are average accuracy over 1800 tasks generated from testing set. After $\pm$ is 95\% confidence intervals over testing tasks. The method with best-performing of ours and prior works are highlighted, and '-' means no report.}
\begin{center}
\setlength{\tabcolsep}{5mm}{
\begin{tabular}{lcccc}
\toprule[1pt]
\textbf{Model} & \multicolumn{2}{c}{\textbf{5-way Acc}} & \multicolumn{2}{c}{\textbf{20-way Acc}} \\
     & 1-shot & 5-shot & 1-shot &5-shot \\
\hline
MANN\cite{Santoro} & 82.8\% & 94.9\% & - & - \\
Siamese nets\cite{Koch} & 97.3\% & 98.4\% & 88.1\% & 97.0\% \\
Matching nets\cite{Oriol} & 98.1\% & 98.9\% & 93.8\% & 98.5\% \\
Neural statistician\cite{Edwards} & 98.1\% & 99.5\% & 93.2\% & 98.1\%\\
Prototypical nets\cite{Jake} & 98.8\% & 99.7\% & 96.0\% & 98.9\%\\
MAML\cite{Chelsea} & 98.7$\pm$0.4\% & 99.9$\pm$0.1\% & 95.8$\pm$0.3\% & 98.9$\pm$0.2\%\\
Meta-SGD\cite{Fengwei} & 99.53$\pm$0.26\% & \textbf{99.93$\pm$0.09}\% & 95.93$\pm$0.38\% & 98.97$\pm$0.19\%\\
Relation nets\cite{Flood} & \textbf{99.6$\pm$0.2\%} & 99.8$\pm$0.1\% & 97.6$\pm$0.2\% & 99.1$\pm$0.1\%\\
SNAIL\cite{Mishra} & 99.07$\pm$0.16\% & 99.78$\pm$0.09\% & 97.64$\pm$0.30\% & 99.36$\pm$0.18\%\\
Support nets4\cite{Jinchao} & 99.24$\pm$0.14\% & 99.75$\pm$0.15\% & 97.79$\pm$0.06\% & 99.27$\pm$0.15\%\\
Support nets6\cite{Jinchao} & 99.37$\pm$0.09\% & 99.80$\pm$0.03\% & \textbf{98.58$\pm$0.07\%} & \textbf{99.45$\pm$0.04\%}\\
\hline
DCN4 & \textbf{99.800$\pm$0.050\%} & 99.891$\pm$0.008\% & \textbf{98.825$\pm$0.025\%} & \textbf{99.505$\pm$0.004\%}\\
\bottomrule[1pt]
\end{tabular}
}
\end{center}
\label{OmniglotResult}
\end{table*}

\subsubsection{MiniImageNet}
The feature embedding was trained using DenseNet-161~\cite{Gao} with 96 initial channels and 16 growth rate. After DenseNet-161 one $1\times1$ convolution layer change the number of channels to 256 without pooling as feature embedding, the feature embedding is not fine-tuned during the ``inner loop''. We use the standard data augmentation from ImageNet:

\begin{itemize}
\vspace{3mm}
 \item random horizontal flipping.
     \vspace{3mm}
 \item resize image into $100\times100$ frame and crop the image to random size and aspect ratio, then resize to $84\times84$.
     \vspace{3mm}
 \item randomly jitter the brightness contrast and saturation of the image.
     \vspace{3mm}
\end{itemize}

After the feature embedding, the structure of resnet block $(3,3)$ include a layer with DCN, a convolution layer with batch normalization and the non-linear function ReLU. Although there is only one layer within DCN, it is still effective because the decoders can be been reused within the layer. The features after global average pooling fed into a fully connected layer with softmax. Models are trained for 40K iterations by Adam with AMSGrad~\cite{Duchi}. The initial learning rate is $10^{-1}$, and decays by 0.5 for every 10K episodes, after 20k episodes we use learning rate cyclic annealing, which is given by:
\begin{equation}
\alpha(t)=\frac{\alpha_0}{2}\left(\cos\left(\frac{\pi{\rm mod}(t-1, T)}{T}\right)+1\right),
\end{equation}
where $\alpha_0$ denotes the initial learning rate, and $T$ is the period of cyclic annealing. We set $T=2000$, and decay $\alpha_0$ by 0.5 for each 10K episodes.

Same as Omniglot model was trained using cross-entropy loss, and use learnable inner learning and weight decay rate, and their learning rate is 0.1 times of other parameters. It is different to Omniglot we do clips of gradient norm of an iterable of parameters during ``inner loop''. After testing on the validation set, we retrain the model on dataset involving training set and validation set with the same hyper-parameter and choose a model with same number of iteration to the ensemble, then testing on the testing set. Since we have not computing power to train a WRN-28-10~\cite{zagoruyko2016wide} by Population-Based Training (PBT)~\cite{jaderberg2017population} like in \cite{Andrei}, we replace DCN by LEO in miniImageNet model same as Omniglot. There are more than 80 million parameters in minImageNet model with LEO, which is much larger than those in the DCN model, which has 4 million parameters.

We evaluate our model on 1-shot, 5-shot, and 5-way tasks. Both of two tasks have 8 tasks in a mini-batch. Each task contains 15 examples as a query. We random generate 1000 tasks through validation and testing set respectively after training.

\subsection{Results}
The results of sinusoid curve fitting are summarised in Table~\ref{SineResult}, which shows that our model has better performance than MAML. It proves that sinusoid function can be learned quickly by DCN. The performance improvement of DCN is more significant when the number of shot is smaller. Compare 5-shot to 20-shot, while the loss of MAML has increased by 28 times, the loss of DCN has only increased by 6 times. This shows that our model has made better use of a small amount of data in the regression task.

The results of Omniglot experiment are listed in Table~\ref{OmniglotLEO} and Table~\ref{OmniglotResult}. The support nets6 in Table~\ref{OmniglotResult} involves 6-layer embedding CNNs, which is different from others. But we involve it here in order to make the table more obvious. Except for 5-way 1-shot task, all our models get higher accuracy than the state-of-the-art. The results of DCN4 in 20-way 1-shot and 5-shot are even better than support nets6, which have more convolutional layers. Besides, the results in Table~\ref{OmniglotLEO} show that DCN has better performance on Omniglot dataset than LEO.

The results of miniImageNet experiment are shown in Table~\ref{MiniImageNetResult}. From this table, most of the large scale models with need co-training or pre-training, but our method obtains the state-of-art results on 5-way 5-shot classification, and comparable results with state-of-art on 5-way 1-shot classification without co-training or pre-training. The comparison between DCN and LEO shows DCN and LEO have similar performance on miniImageNet. However, DCN has much fewer parameters than LEO.

Improvements from ensemble can be observed in Table~\ref{OmniglotLEO} and Table~\ref{MiniImageNetResult}. Ensemble method enhances most of the results, and it is more significant on miniImageNet, which has higher generation gaps.

\renewcommand\arraystretch{1.5}
\begin{table*}
\small
\caption{\upshape The results of few-shot classification on miniImageNet dataset, which are average accuracy over 1000 tasks generated from test dataset. After $\pm$ is 95\% confidence intervals over testing tasks. The first set is the results use convolutional networks, and the results of using much deeper network with resnet or dense net block are showed in the second layer. The method with best-performing of ours and prior works are highlighted.}
\begin{center}
\setlength{\tabcolsep}{5mm}{
\begin{tabular}{lcccc}
\toprule[1pt]
\textbf{Model} & \textbf{Fine Tune} & \textbf{Co- or Pre-training} & \multicolumn{2}{c}{\textbf{5-way Acc}} \\
   &  &  & 1-shot & 5-shot \\
\hline
Matching nets\cite{Oriol} & N & N & 43.56$\pm$0.84\% & 55.31$\pm$0.73\% \\
Meta-learner LSTM\cite{Sachin} & N & N & 43.44$\pm$0.77\% & 60.60$\pm$0.71\% \\
MAML\cite{Chelsea} & Y & N & 48.70$\pm$1.84\% & 63.11$\pm$0.92\% \\
Prototypical nets\cite{Jake} & N & N & 49.42$\pm$0.78\% & 68.20$\pm$0.66\% \\
Meta-SGD\cite{Fengwei} & Y & N & 50.47$\pm$1.87\% & 64.03$\pm$0.94\% \\
Reptile\cite{Alex} & Y & N & 49.97$\pm$0.32\% & 65.99$\pm$0.58\% \\
Relation nets\cite{Flood} & N & N & 50.33$\pm$0.82\% & 65.32$\pm$0.70\% \\
Support nets\cite{Jinchao} & N & N & 56.32$\pm$0.47\% & 71.94$\pm$0.37\% \\
\hline
SNAIL\cite{Mishra} & N & N & 55.71$\pm$0.99\% & 68.88$\pm$0.92\% \\
Dynamic Few-Shot Visual Learning \cite{Spyros} & Y & Y & 56.20$\pm$0.86\% & 73.00$\pm$0.64\% \\
Discriminative k-shot learning \cite{Bauer} & Y & Y & 56.30$\pm$0.40\% & 73.90$\pm$0.30\% \\
Predicting Parameters from Activations \cite{Siyuan} & Y & Y & 59.60$\pm$0.41\% & 73.74$\pm$0.19\% \\
TADAM \cite{Oreshkin} & N & Y & 58.50$\pm$0.30\% &76.70$\pm$0.30\% \\
LEO \cite{Andrei} & Y & Y & \textbf{61.76$\pm$0.08\%} & \textbf{77.59$\pm$0.12\%} \\
\hline
LEO (training) & Y & N & 56.42$\pm$0.08\% & 72.94$\pm$0.07\% \\
LEO-E (training) & Y & N & 57.57$\pm$0.08\% & 77.28$\pm$0.06\% \\
LEO (training plus validation) & Y & N & 57.57$\pm$0.08\% & 74.38$\pm$0.06\% \\
LEO-E (training plus validation) & Y & N & 57.52$\pm$0.08\% & \textbf{78.04$\pm$0.06\%} \\
\hline
DCN (training) & Y & N & 56.72$\pm$0.09\% & 72.39$\pm$0.07\% \\
DCN-E (training) & Y & N & \textbf{58.94$\pm$0.08\%} & 77.13$\pm$0.06\% \\
DCN (training plus validation) & Y & N & 57.09$\pm$0.08\% & 73.48$\pm$0.07\% \\
DCN-E (training plus validation) & Y & N & 58.73$\pm$0.08\% &77.93$\pm$0.06\% \\
%FPMR & Y & N & 55.77$\pm$0.08\% & 73.79$\pm$0.07\% \\
%FPMR-E & Y & N & \textbf{59.00$\pm$0.08\%} & \textbf{77.03$\pm$0.06\%} \\
\bottomrule[1pt]
\end{tabular}
}
\end{center}
\label{MiniImageNetResult}
\end{table*}

\subsection{Discussion}
The proposed approach benefits from DCN, the feature embedding, and ensemble learning. First, DCN improves the performance of the model by updating parameters in the low-dimension space. However, even if we adopt DCN, it did not increase the size of the model, and even reduced it. The reasons are the model parameters of the layer with DCN is replaced by latent code which is a much lower dimension and DCN has fewer parameters itself.

Second, the feature embedding works well, it is effective and efficient. Fine-tuning is only required on last few layers, while the feature embedding can be updated during the ``outer loop'' to enhances the evaluation results of the ``outer loop''. This enables the learning of useful features to meta-learning, and gets higher performance than many models with the pre-training feature embedding. This method makes the training of training large-scale meta-model end-to-end.

Finally, the results show the performance improvement of ensemble learning on miniImageNet is more significant than that on Omniglot. It is same as we speculate that the task with the more larger generation gap between the training set and validation set is more suitable for snapshot ensemble because it is able to get the models with higher quality and diversity.

\section{Conclusion}\label{Sec_Conclusion}
This paper proposes a meta-model using DCN, the results show that DCN is able to denote task-general information. In addition, ensemble learning is integrated with DCN to improve model generalization ability. The experimental results on a benchmark dataset show the effectiveness of solving few-shot learning problems.

In future work, despite of promising better experimental results, the work can be improved by replacing the decoder networks with other kinds of neural networks. In addition, DCN may be applied to reinforcement learning tasks for quick adaptation; learning environment information by the low-dimension parameter space may make deep reinforcement learning model more stable during the training.

\begin{appendices}
\section{}
\subsection{Task feature}\label{Taskfeatkernel}

\begin{figure*}
\begin{center}
\includegraphics[width=150mm]{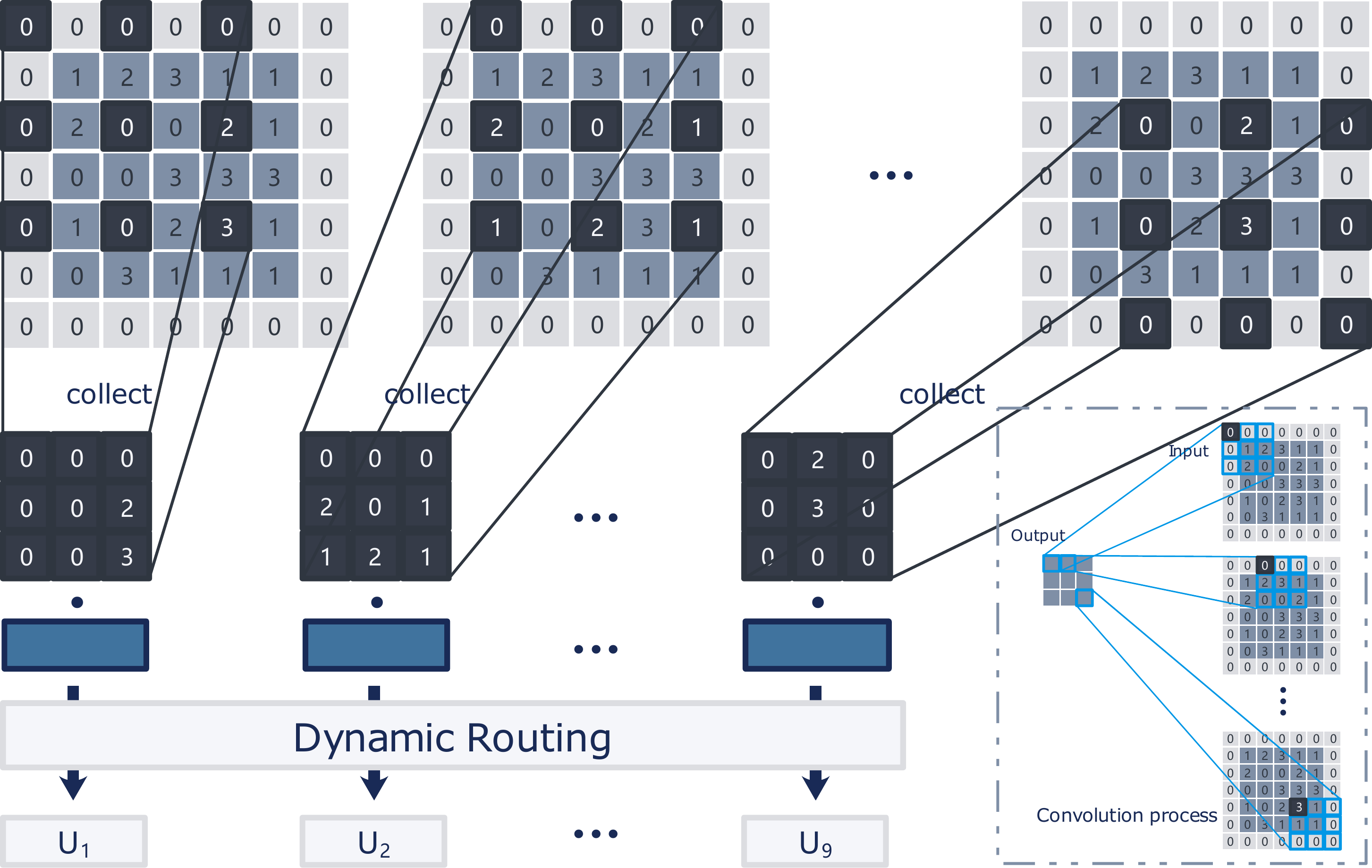}
    \put(-407,66){\small \color{LYNX WHITE} $\rm w$}
    \put(-313,66){\small \color{LYNX WHITE} $\rm w$}
    \put(-176,66){\small \color{LYNX WHITE} $\rm w$}
\end{center}
    \caption{An example of kernel size $9\times9$ and $C_{in}=1$, where ${\rm U}_l$ is the output capsule. The kernel size, stride and zero-padding and a channel size of the layer are $9\times9$, $2\times2$, $1\times1$ and 1 respectively. When channel size is not equal to 1, ${\rm w}$ is not a vector but a matrix, but the number of ${\rm U}_l$ remains unchanged. The lower right corner of the graph is the operation process after the generated parameters. The convolution operation is consistent with standard CNN.}
\label{CapsuleMatch}
\end{figure*}

%The features used to calculate firing strengths comes from the input features of this layer.
Assume that the size of the input ${\rm u}$ of the network layer is $(N_d,C_{in},W_{in},H_{in})$, where $N_d$ and $C_{in}$ are the numbers of examples and channels respectively, and $W_{in}$ and $H_{in}$ are the width and height respectively. The output size is $(N_d,C_{ou},W_{ou},H_{ou})$. %{\color{brown}The $w_1\times w_2$ weight matrix of one dimension of kernel is only related to the features do multiply operation with, where $w_1$ and $w_2$ are the number of input and output channels.}
One example of the features choice is shown in Fig.~\ref{CapsuleMatch}. In Fig.~\ref{CapsuleMatch} kernel size is $9\times9$ and $C_{in}=1$. The mathematical form of channel $e$ is given by,
\begin{equation}
\label{eq_u_matrix}
\begin{aligned}
{\rm u}^{ne}_k=&\left[
    \begin{array}{cccc}
        u^{ne}_{h^k_1,w^k_1} & u^{ne}_{h^k_1,w^k_2} & \cdots & u^{ne}_{h^k_1,w^k_{W_{ou}}}\\
        u^{ne}_{h^k_2,w^k_1} & u^{ne}_{h^k_2,w^k_2} & \cdots & u^{ne}_{h^k_2,w^k_{W_{ou}}}\\
        \vdots & \vdots & \ddots & \vdots\\
        u^{ne}_{h^k_{H_{ou}},w^k_1} & u^{ne}_{h^k_{H_{ou}},w^k_2} &\cdots & u^{ne}_{h^k_{H_{ou}},w^k_{W_{ou}}}
    \end{array}\right]\\
&(n=1,2,\cdots,N_d),(e=1,2,\cdots,C_{in}),
\end{aligned},
\end{equation}
\begin{equation}
\label{eq_u_vec}
    {\rm u}^{e}_k={\rm vec}\left(\left[{\rm u}^{1e}_k,{\rm u}^{2e}_k,\cdots,{\rm u}^{Ne}_k\right]\right),
\end{equation}
where $e$ and $k$ are the indexes of channels and kernel dimensions, respectively; $h^k_i=h^k_0+(i-1)str_1,\ (i=1,2,\cdots,H_{ou})$ and $w^k_i=w^k_0+(i-1)str_2,\ (i=1,2,\cdots,W_{ou})$, $h^k_0$ and $w^k_0$ are the starting coordinates of scanning; $str_1$ and $str_2$ are the strides in row and column respectively, and $\rm vec()$ is the operation of pulling the matrix into a row vector. The weights of a capsule layer are shared by the same channel cross different examples. Last, the task features are received by the capsule layer is given by:

\begin{equation}
\label{eq_wu}
\begin{aligned}
    \hat{\rm u}_k&={\rm cat}\left({\rm w}^1_k{\rm u}^{1}_k,{\rm w}^2_k{\rm u}^{2}_k,\cdots,{\rm w}^{C_{in}}_k{\rm u}^{C_{in}}_k\right)\\
    &=\left[
    \begin{array}{cccc}
    \hat{u}_{1|1} & \hat{u}_{1|2} & \cdots & \hat{u}_{1|J}\\
    \hat{u}_{2|1} & \hat{u}_{2|2} & \cdots & \hat{u}_{2|J}\\
    \vdots & \vdots & \ddots & \vdots\\
    \hat{u}_{N_f|1} & \hat{u}_{N_f|2} & \cdots & \hat{u}_{N_f|J}
    \end{array}
    \right],
\end{aligned}
\end{equation}
where $\rm cat()$ denotes the concatenation of matrix, $N_f$ is the number of state variables, $J=H_{ou}W_{ou}N_dC_{in}$, and ${\rm w}^e_k$ is a column vector with length $N_f$.
%We divide each state variable into two fuzzy sets, so here $S=2^K$.

\subsection{\color{blue}Dynamic routing}\label{State_variables}
After getting the features, {\color{blue}dynamic routing}~\cite{Sara} is adopted to calculate the output capsule. The `prediction vectors''~\cite{Sara} is given by:
\begin{equation}
\label{eq_ui}
\hat{\rm u}_{\cdot|j}=\left[\hat{u}_{1|j},\hat{u}_{2|j},\cdots,\hat{u}_{N_f|j}\right]^{\rm T},
\end{equation}
where $j=1,2,\cdots,J$. Each $\hat{\rm u}_{\cdot|j}$ accumulates according to coupling coefficients and non-linear squashing is used to shrink the module of vector to $0\sim1$:
\begin{equation}
\label{eq_s}
\hat{\rm s}=\sum_j{\rm L}_{j}\circ\hat{\rm u}_{\cdot|j},
\end{equation}
\begin{equation}
\label{eq_squashing}
{\rm v}=\frac{\|\hat{\rm s}\|^2}{1+\|\hat{\rm s}\|^2}\frac{\hat{\rm s}}{\|\hat{\rm s}\|},
\end{equation}
where $\circ$ denotes element-wise multiplication. In this case, the input includes $J$ capsule the length of which is 1, and the output includes 1 capsule the length of which is $N_f$; the squashing operation is applied to the entire output $\hat{\rm s}$. ${\rm L}_j$ is the coupling coefficients that are calculated by iterative dynamic routing~\cite{Sara} through:
\begin{equation}
\label{eq_softmax}
L_{ij}=\frac{\exp(b_{ij})}{\sum_k\exp(b_{kj})},
\end{equation}
where ${\rm L}_j=\{L_{ij}\}_i$, and ${\rm b}=\{b_{ij}\}_{ij}$ denotes the correlation between $\hat{\rm u}_l$ and {\rm v}. The initial value of $\rm b$ is a zero matrix. After getting the output $\rm v$, $\rm b$ will be recalculated by:
\begin{equation}
\label{eq_b}
{\rm b}_{n\cdot}=\hat{\rm u}_{n|\cdot}v_n,\ (n=1,2,\cdots,N_f),
\end{equation}
where ${\rm b}_{n\cdot}$ and $\hat{\rm u}_{n|\cdot}$ are row vectors of ${\rm b}$ and $\hat{\rm u}_k$ respectively, and $v_n$ is the element of $\rm v$. Repeat the process based on Eq. $(\ref{eq_softmax})\rightarrow(\ref{eq_s})\rightarrow(\ref{eq_squashing})\rightarrow(\ref{eq_b})$ $r$ times after initialising $\rm b$. Following the work of \cite{Sara} $r=3$ is applied in all the experiments.

\end{appendices}

% if have a single appendix:
%\appendix[Proof of the Zonklar Equations]
% or
%\appendix  % for no appendix heading
% do not use \section anymore after \appendix, only \section*
% is possibly needed

% use appendices with more than one appendix
% then use \section to start each appendix
% you must declare a \section before using any
% \subsection or using \label (\appendices by itself
% starts a section numbered zero.)
%

\appendices

% you can choose not to have a title for an appendix
% if you want by leaving the argument blank

% use section* for acknowledgment

% Can use something like this to put references on a page
% by themselves when using endfloat and the captionsoff option.
\ifCLASSOPTIONcaptionsoff
  \newpage
\fi

% trigger a \newpage just before the given reference
% number - used to balance the columns on the last page
% adjust value as needed - may need to be readjusted if
% the document is modified later
%\IEEEtriggeratref{8}
% The "triggered" command can be changed if desired:
%\IEEEtriggercmd{\enlargethispage{-5in}}

% references section

% can use a bibliography generated by BibTeX as a .bbl file
% BibTeX documentation can be easily obtained at:
% http://mirror.ctan.org/biblio/bibtex/contrib/doc/
% The IEEEtran BibTeX style support page is at:
% http://www.michaelshell.org/tex/ieeetran/bibtex/
%\bibliographystyle{IEEEtran}
% argument is your BibTeX string definitions and bibliography database(s)
%\bibliography{IEEEabrv,../bib/paper}
%
% <OR> manually copy in the resultant .bbl file
% set second argument of \begin to the number of references
% (used to reserve space for the reference number labels box)

\bibliographystyle{IEEEtran}
\bibliography{paper}

\end{document}